%% file: main.tex
\documentclass[10pt,twocolumn,letterpaper]{article}

\usepackage[pagenumbers]{cvpr}


\definecolor{cvprblue}{rgb}{0.21,0.49,0.74}
\definecolor{generalization}{RGB}{100, 130, 253}
\definecolor{memorization}{RGB}{243, 165, 57}
\usepackage[hyphens]{url}
\usepackage[pagebackref,breaklinks,colorlinks,allcolors=cvprblue]{hyperref}
\usepackage{mathtools}
\usepackage{booktabs, multirow}
\usepackage{amsthm}
\theoremstyle{plain}
\newtheorem{theorem}{Theorem}[section]
\usepackage[table]{xcolor}
\usepackage{stfloats}
\usepackage[accsupp]{axessibility}  

\usepackage{tcolorbox}
\newtcolorbox{findingbox}{
  colback=gray!10, colframe=black!80,
  boxrule=1pt, arc=4pt,
  left=6pt, right=6pt, top=6pt, bottom=6pt,
  before skip=6pt, after skip=6pt,
}

\def\paperID{}
\def\confName{CVPR}
\def\confYear{2026}

\title{From Navigation to Refinement: Revealing the Two-Stage Nature of\\Flow-based Diffusion Models through Oracle Velocity}

\author{Haoming Liu$^{1,2}$ $\quad$ Jinnuo Liu$^{1}$ $\quad$ Yanhao Li$^{1}$ $\quad$ Liuyang Bai$^{1}$ $\quad$ Yunkai Ji$^{1}$ $\quad$ Yuanhe Guo$^{1,2}$ \\ \\
Shenji Wan$^{1,2}$ $\quad$ Hongyi Wen$^{1,2}$ \\ \\
$^1$Center for Data Science, New York University Shanghai $\quad$ $^2$New York University
}

\begin{document}
\maketitle
\input{sec/0_abstract}    
\input{sec/1_intro}

\input{sec/2_method}
\input{sec/3_experiment}
\input{sec/4_related_work}

\input{sec/5_conclusion}
\newpage

\section*{Acknowledgement}
This work is supported by NYU Shanghai Center for Data Science and SMEC DFYC fund. This work is supported in part through the NYU IT High Performance Computing resources, services, and staff expertise.

{
    \small
    \bibliographystyle{ieeenat_fullname}
    \bibliography{main}
}

\input{sec/X_suppl}

\end{document}

%% file: sec/0_abstract.tex
\begin{abstract}
Flow-based diffusion models have emerged as a leading paradigm for training generative models across images and videos. However, their memorization-generalization behavior remains poorly understood.
In this work, we revisit the flow matching (FM) objective and study its marginal velocity field, which admits a closed-form expression, allowing exact computation of the oracle FM target.
Analyzing this oracle velocity field reveals that flow-based diffusion models inherently formulate a two-stage training target: an early stage guided by a mixture of data modes, and a later stage dominated by the nearest data sample.
The two-stage objective leads to distinct learning behaviors: the early navigation stage generalizes across data modes to form global layouts, whereas the later refinement stage increasingly memorizes fine-grained details.
Leveraging these insights, we explain the effectiveness of practical techniques such as timestep-shifted schedules, classifier-free guidance intervals, and latent space design choices.
Our study deepens the understanding of diffusion model training dynamics and offers principles for guiding future architectural and algorithmic improvements.
Our project page is available at: \url{https://maps-research.github.io/from-navigation-to-refinement/}.
\end{abstract}

%% file: sec/1_intro.tex
\section{Introduction}
\label{sec:intro}

Diffusion models~\cite{sohl2015deep, ho2020denoising, dhariwal2021diffusion, song2021ddim} have emerged as a powerful class of generative methods, capable of synthesizing high-fidelity samples across diverse domains, such as images~\cite{podell2023sdxl, esser2024sd3, flux2023, labs2025flux, nano_banana} and videos~\cite{sora_openai_2024, peng2025opensora, blattmann2023svd, wan2025wan}. These models learn complex data distributions by progressively transforming a prior distribution (e.g., Gaussian) into the data distribution through an interpolation process parameterized by denoising or transport directions. Various formulations have been proposed to characterize the diffusion process, including probabilistic models~\cite{ho2020denoising, song2021ddim}, score-based ODEs/SDEs~\cite{song2019generative, song2020improved, song2020score}, and flow matching~\cite{albergo2023interpolant, liu2023flow, lipman2023flow}. These perspectives give rise to diverse training objectives (e.g., noise, score, or velocity prediction) and correspond to different parameterizations of the probability flow ODE (PF-ODE)~\cite{song2020score, lai2025principles}. Recent advances have largely converged on the flow matching formulation, where the model is trained to predict the velocity field under a linear schedule (also known as the canonical linear flow or rectified flow~\cite{liu2023flow}). Thanks to its simplicity and stable training dynamics, this canonical formulation has become the de facto standard for training state-of-the-art diffusion models.

\begin{figure}[t!]
\centering
\includegraphics[width=\linewidth]{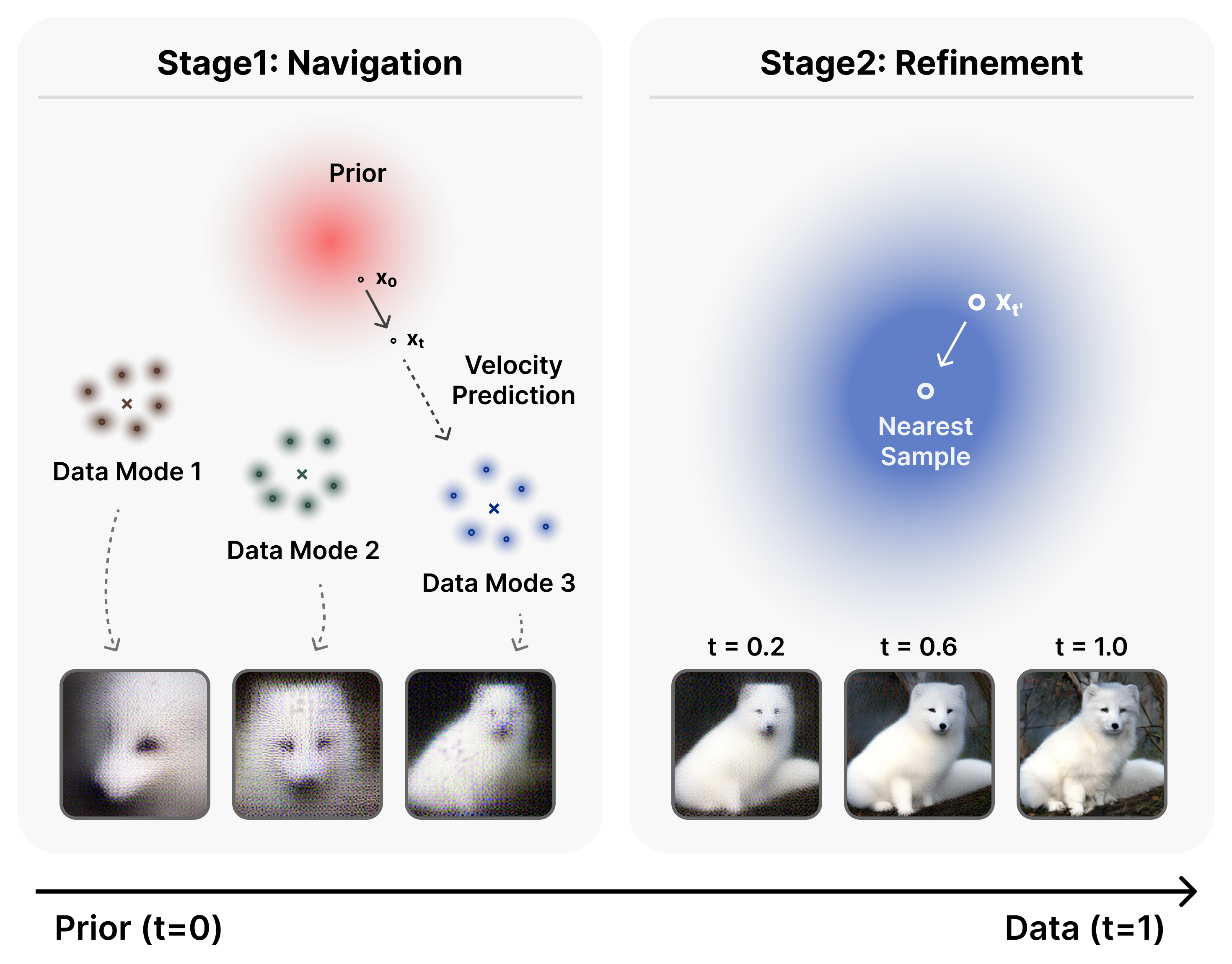}
\vspace{-8pt}
\caption{Illustration of the two stages in flow-based diffusion models. In the navigation stage (near prior), the target is guided by a mixture of multiple data samples, forming global layouts. In the refinement stage (near data), the target is dominated by the nearest data sample, refining fine-grained visual details.}
\label{fig:teaser}
\vspace{-8pt}
\end{figure}

Meanwhile, a growing line of research has sought to understand the training and inference behaviors of diffusion models, particularly on the balance between \textcolor{memorization}{memorization} (i.e., the tendency to reproduce training samples) and \textcolor{generalization}{generalization} (i.e., the ability to synthesize novel samples). Earlier works analyze this phenomenon from diverse viewpoints, such as by deriving quantitative metrics to characterize model behavior~\cite{gu2024on, zhang2025understanding} and by analyzing the underlying factors that lead to memorization/generalization~\cite{zhang2023emergence, kadkhodaie2024generalization, shi2025a}. In particular, most works primarily: (1) focus on the model behaviors when sampling from scratch, and (2) verify their findings on datasets of small scales/resolutions (e.g., FFHQ~\cite{karras2019style}, CIFAR-10~\cite{krizhevsky2010cifar}). While such insights can be informative in low-data regimes, such memorization behaviors become increasingly unlikely as the data scales to ImageNet-level~\cite{deng2009imagenet} or beyond. Song \textit{et al.}~\cite{song2025selective} recently observe the memorization/generalization divergence on ImageNet-scale diffusion models when resuming sampling from different temporal ranges. In general, resuming from an earlier timestep (near the prior) leads to novel samples, whereas later ones tend to reproduce training images. Hence, a key question arises: \textit{what underlying principles govern the balance between \textcolor{memorization}{memorization} and \textcolor{generalization}{generalization} in diffusion models trained on large-scale data?}

To address this, we trace the origin of model behavior back to its training objective. We first revisit flow matching (FM)~\cite{lipman2023flow, liu2023flow} and its gradient-equivalent proxy, conditional flow matching (CFM), where the former is generally regarded as intractable due to the unavailable ground-truth velocity field. In this work, we refine this conventional view by showing that the marginal velocity field of rectified flow admits a closed-form expression under a Gaussian prior and a finite training set (Sec.~\ref{sec:oracle}). This allows us to compute the \textbf{oracle velocity} (defined by the FM objective) at any location in the sample space. Through the lens of the oracle velocity field, the effective training target of flow-based diffusion models exhibits two distinctive stages (Sec.~\ref{sec:two_stage}). In the early stage that is closer to the prior (termed the \textcolor{generalization}{navigation} stage), the oracle velocity contains combined information of multiple data points, and guides the model toward a mixture of relevant data modes. In the later stage (termed the \textcolor{memorization}{refinement} stage), the velocity field is dominated by a single data point. In addition, we identify \textit{data dimensionality} and \textit{sample size} as the two key factors for determining the point of stage transition over time.

We next analyze the model’s behavior in light of this two-stage structure (Sec.~\ref{sec:model_behavior}).
Overall, the navigation stage primarily establishes the global image layout, whereas the refinement stage concentrates on polishing fine-grained visual details.
We hypothesize that the model’s \textcolor{generalization}{generalization} capability arises mainly from the \textcolor{generalization}{navigation} stage, while its \textcolor{memorization}{memorization} behavior stems from the \textcolor{memorization}{refinement} stage.
Moreover, we observe that the learning difficulty differs across stages: the refinement stage poses a greater challenge to learn, while the navigation stage is comparatively easier.

We leverage our stage-level insights to explain why several empirically effective techniques succeed (Sec.~\ref{sec:implications}), including: (1)~timestep shifting for non-uniform sampling schedules, (2)~classifier-free guidance (CFG) interval selection, and (3)~latent space design choices. We further discuss how these findings inform broader practices and highlight underexplored directions for future improvement.

%% file: sec/2_method.tex
\section{Oracle Velocity from Empirical Mixture}
\label{sec:oracle}

\subsection{Preliminaries on Flow Matching}
We consider two endpoint distributions: a simple prior distribution \( p_{\text{prior}} \) serving as the source, and a target data distribution \( p_{\text{data}} \) as the destination. The goal is to learn a continuous transport map that evolves the source into the target. This process induces a family of intermediate marginal densities \( \{p_t\}_{t\in[0,1]} \) defined over a normalized time interval \( t \in [0,1] \)~\footnote{The time conventions used in diffusion and flow matching works are highly inconsistent; here we follow Lipman \textit{et al.}~\cite{lipman2023flow}, letting \( t=0 \) corresponds to the prior distribution and \( t=1 \) to the data distribution.}. Despite the formulation differences between diffusion models and flow matching, their intrinsic dynamics are both governed by the \emph{continuity equation}:
\begin{equation}
\frac{\partial p_t(x)}{\partial t} + \nabla_x \!\cdot\! (u_t(x) p_t(x)) = 0,
\end{equation}
which couples the evolving probability path \( p_t \) with the underlying velocity field \( u_t(x) \) that transports the probability mass along the path. The flow matching (FM) objective learns a neural network $v_t(x_t; \theta)$ to regress $u_t(x_t)$:
\begin{equation}
    \mathcal{L}_{\text{FM}}(\theta) = \mathbb{E}_{t, \, p_t(x_t)} || v_t(x_t; \theta) - u_t(x_t) ||^2.
    \label{eq:fm}
\end{equation}
However, the FM objective is generally intractable when a closed-form $u_t$ is unavailable, so a common practice is to construct a conditional probability path $p_t(x_t \mid x_1)$ based on a particular data sample $x_1$. This is also known as the conditional flow matching (CFM) objective:
\begin{equation}
    \mathcal{L}_{\text{CFM}}(\theta) = \mathbb{E}_{t, \,q(x_1), \, p_t(x_t \mid x_1)} || v_t(x_t; \theta) - u_t(x_t \mid x_1) ||^2,
    \label{eq:cfm}
\end{equation}
which has been shown to share identical gradients with the FM objective. Meanwhile, the probability path $p_t$ can be constructed in various ways. One predominant approach is to randomly sample $x_0 \sim p_{\text{prior}}$, $x_1 \sim p_{\text{data}}$, and interpolate linearly via time-dependent scaling factors $\alpha_t$ and $\sigma_t$:
\begin{equation}
x_t = \alpha_t x_1 + \sigma_t x_0.
\label{eq:interpolation}
\end{equation}
Hence, the conditional training target is given by:
\begin{equation}
u_t(x_t \mid x_1) = \dot \alpha_t x_1 + \dot \sigma_t x_0.
\label{eq:cond_ut}
\end{equation}
For rectified flow, we set $(\alpha_t, \sigma_t) = (t, 1-t)$ and adopt $u_t(x_t\mid x_1) = x_1 - x_0$ as the training target.

\subsection{Closed-Form Oracle under Gaussian Prior}
\label{sec:oracle_closedform}

While the original Flow Matching (FM) objective (Eq.~\ref{eq:fm}) is intractable under unknown probability paths, it admits a closed-form oracle velocity field under a few mild conditions in the context of flow-based diffusion models.  
Specifically, we assume (1) a Gaussian prior distribution, (2) a finite dataset $\{x_1^{(i)}\}_{i=1}^N$ approximating $p_{\text{data}}$, and (3) a linear interpolation path as in rectified flow. Under these assumptions, we can explicitly compute the conditional expectation $\mathbb{E}[u_t(x_t\mid x_1)\mid x_t]$, which yields the following closed-form expression for the oracle velocity field.

\begin{theorem}[Closed-Form Oracle under Gaussian Prior]
\label{thm:oracle_closedform}
Let the data distribution be represented as an empirical mixture over finite dataset $\{x_1^{(i)}\}_{i=1}^N$ and consider the linear interpolation $x_t = \alpha_t x_1^{(i)} + \sigma_t x_0$ with $x_0 \sim \mathcal{N}(0, I)$ and uniformly sampled $x_1^{(i)}$. Then, the oracle velocity field $u_t^*(x_t,t) \coloneqq \mathbb{E}\!\left[u_t(x_t\mid x_1)\mid x_t\right]$ admits the closed form:
\begin{equation}
  u_t^*(x_t,t)
  \;=\;A_t\sum_{i=1}^N \gamma_i(x_t,t)\,x_1^{(i)}\;+\;B_t\,x_t,
  \label{eq:oracle_closed}
\end{equation}
where the coefficients $A_t = \dot{\alpha}_t - \frac{\alpha_t \dot{\sigma}_t}{\sigma_t},
B_t = \frac{\dot{\sigma}_t}{\sigma_t},$ and the normalized posterior weights $\gamma_i(x_t,t)$ are given by:
\begin{equation}
\gamma_i(x_t,t)\;=\;
\frac{\exp\!\big(-\tfrac{\|x_t-\alpha_t x_1^{(i)}\|^2}{2\sigma_t^2}\big)}
  {\sum_{j=1}^N \exp\!\big(-\tfrac{\|x_t-\alpha_t x_1^{(j)}\|^2}{2\sigma_t^2}\big)}.
\label{eq:gamma_closed}
\end{equation}
\end{theorem}

\noindent
The proof can be given by applying Bayes’ rule to the path marginal (a Gaussian mixture) and taking the conditional expectation given $x_t$; the full derivation is provided in the Appendix. For class-conditional generation, the oracle velocity can be computed within each class-specific subset of samples $\{x_1^{(i)}\}_{i \in \mathcal{I}_y}$, denoted as $u_t^*(x_t,t\mid y)$. We also note that prior work has explored the closed-form targets under various diffusion formulations~\cite{bertrand2025on, kamb2025an, biroli2024dynamical, gao2024flow, li2024good, scarvelis2025closedform}.

\section{The Two-Stage Training Target}
\label{sec:two_stage}

Given the closed-form expression in Eq.~\ref{eq:oracle_closed}, Flow Matching training can be viewed as a supervised learning problem, where the network is tasked with predicting the target label $u_t^*(x_t, t)$ for each input pair $(x_t, t)$.
Interestingly, this (conceptually infinite) training dataset comprises samples with two distinct characteristics, separated by the timestep $t$. 
Roughly speaking, for $x_t$ samples with $ t \in [0.0, 0.1] $, the training target $ u_t^*(x_t, t) $ is influenced by multiple data points from the target distribution, whereas for larger $t$, the target becomes dominated by a single nearest data point.

\begin{figure}[t!]
\centering
\includegraphics[width=0.98\linewidth]{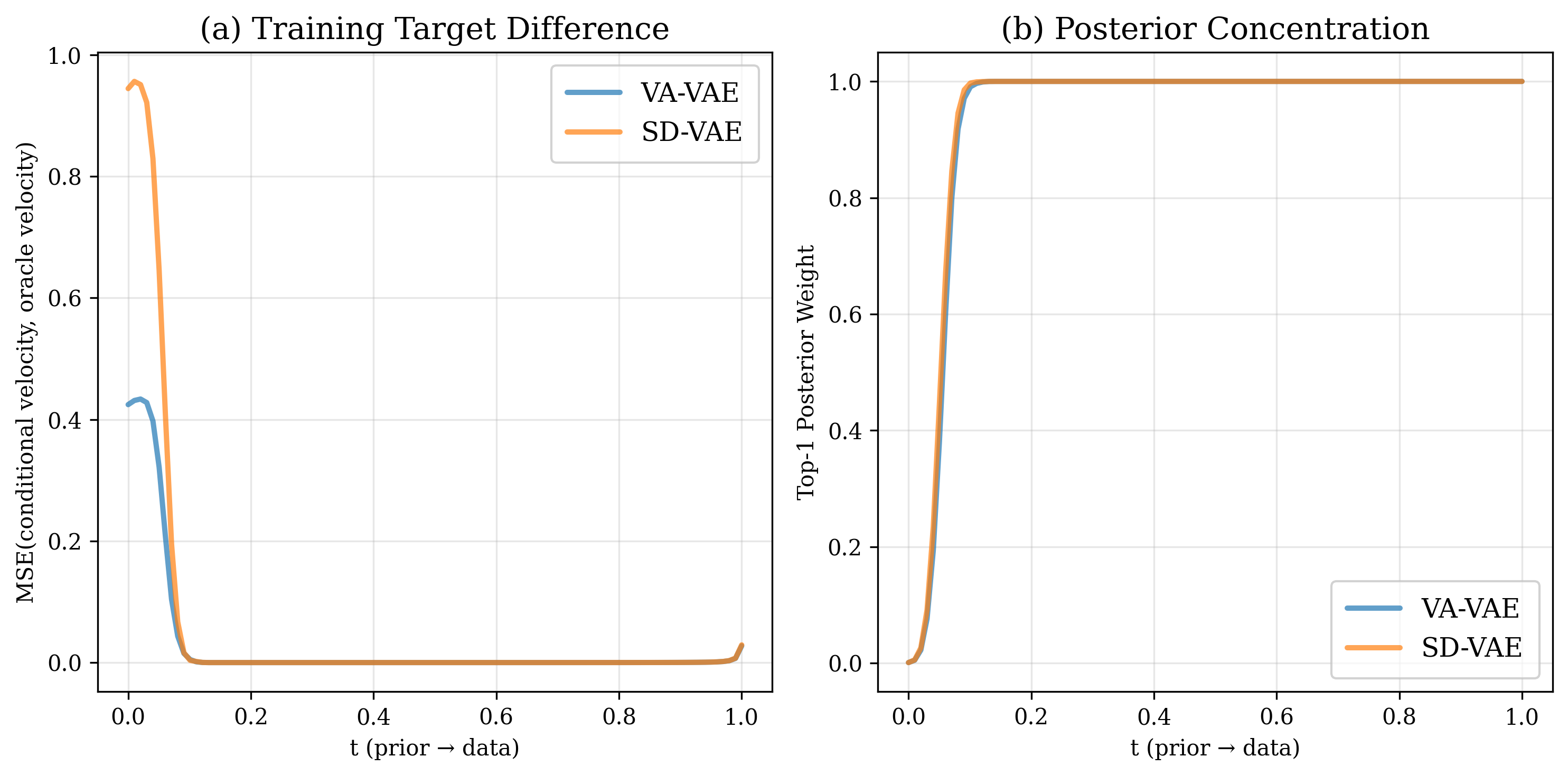}
\vspace{-4pt}
\caption{(a) MSE between $u_t^*$ and the CFM target $(x_1 - x_0)$ across timesteps; (b) Average top-1 posterior weight $\gamma_i(x_t, t)$ showing rapid concentration after $t=0.1$; both plots reveal a clear two-stage behavior emerging in the oracle training target.}
\vspace{-4pt}
\label{fig:two_stage}
\end{figure}

Specifically, we consider a class-conditional image generation setting on the ImageNet~\cite{deng2009imagenet} dataset, where we train flow-based diffusion models within the latent space of VA-VAE~\cite{yao2025reconstruction} and SD-VAE~\cite{rombach2022ldm}. Under rectified flow~\cite{liu2023flow}, the noisy latents $x_t$ are constructed from randomly sampled prior/data pairs $(x_0, x_1)$ following Eq.~\ref{eq:interpolation}. Then, we can compute the class-conditioned oracle velocity $u_t^*(x_t, t \mid y)$ and contrast against the CFM target $(x_1 - x_0)$. As shown in Fig.~\ref{fig:two_stage}(a), the MSE between the noisy and oracle targets reveals that the divergence is concentrated in the early interval $t \in [0.0, 0.1]$, while the two targets closely align for later timesteps ($t > 0.1$). A minor discrepancy is also observed as $t \to 1$, which arises from the rapidly shrinking $2\sigma_t^2$ term in the denominator of $\gamma_i$. Meanwhile, the two-stage target can be further validated through the top-1 posterior weight among $\gamma_i(x_t, t)$ (Fig.~\ref{fig:two_stage}(b)), where the normalized top-1 weight rapidly saturates to 1 beyond $t = 0.1$, suggesting that the oracle velocity field has collapsed to a single dominant data point. Together, these results confirm that the oracle training target inherently exhibits a two-stage nature, with a clear transition around $t \approx 0.1$.

\begin{figure}[t!]
\centering
\includegraphics[width=0.98\linewidth]{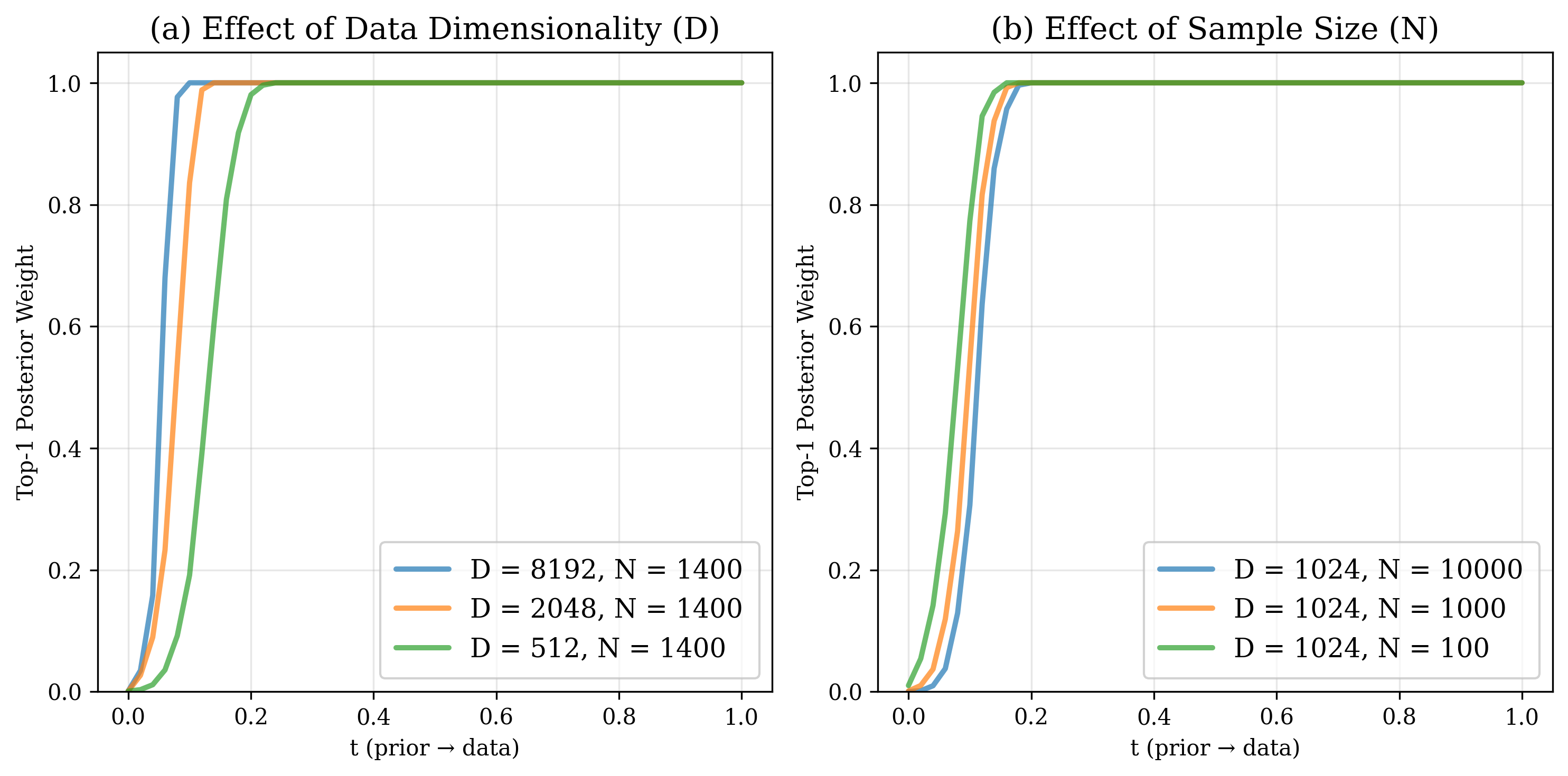}
\vspace{-4pt}
\caption{Plots of top-1 posterior weight under varying conditions. (a) Higher data dimensionality accelerates saturation. (b) A larger sample size delays the transition. Zoom in for the details.}
\label{fig:split_point}
\vspace{-8pt}
\end{figure}

\begin{figure*}[t!]
\centering
\includegraphics[width=0.98\linewidth]{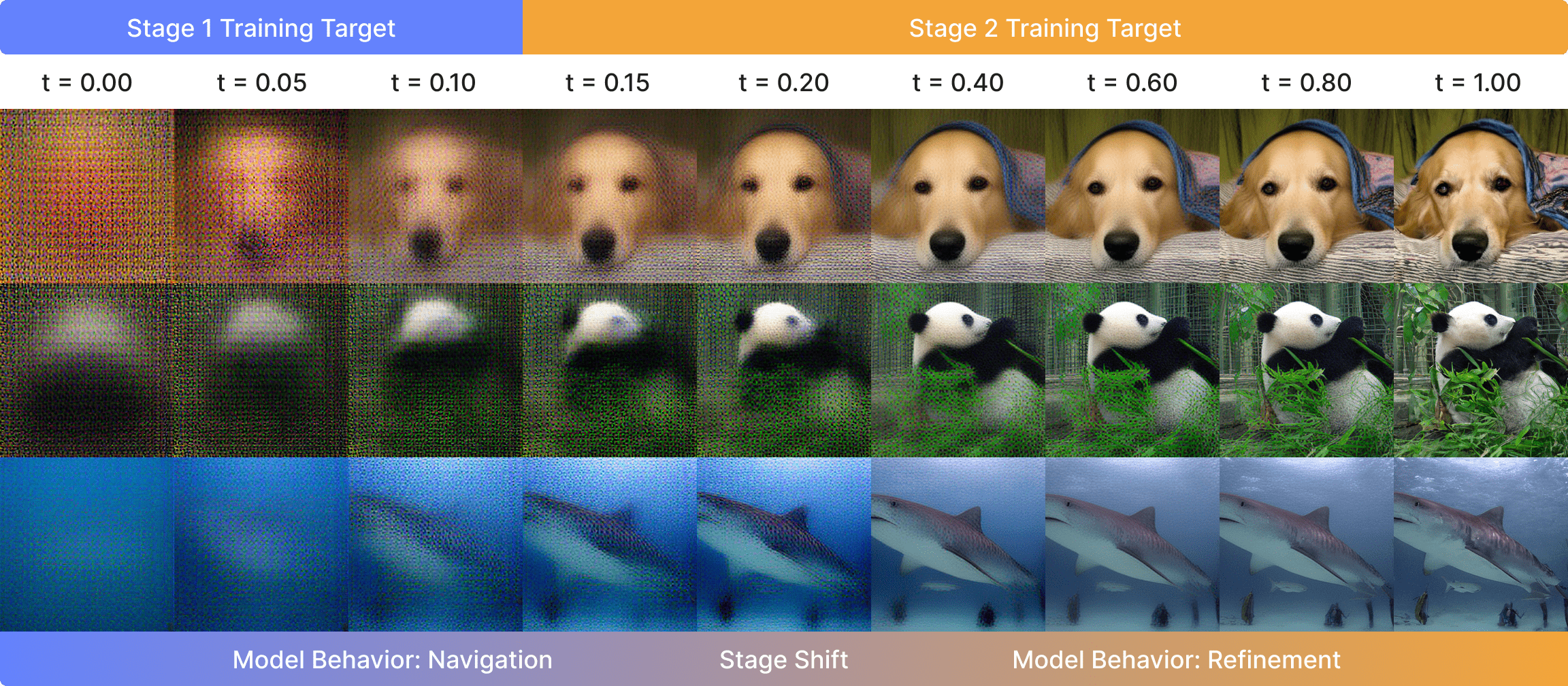}
\caption{Intermediate predictions of a LightningDiT-XL/1~\cite{yao2025reconstruction} model across timesteps. Overall, early stages primarily \textcolor{generalization}{navigate} global layout, while later stages \textcolor{memorization}{refine} fine-grained details. Notably, the empirically observed stage transition ($t\approx0.2$) in model behavior lags slightly behind the training target shift ($t\approx0.1$). Zoom in for the best view. Better view with color.}
\label{fig:qual_model_pred}
\vspace{-4pt}
\end{figure*}

Here, the rapid saturation of the top-1 posterior mass stems from the posterior weighting defined in Eq.~\ref{eq:gamma_closed}. More specifically, we first recall the conditional case (i.e., considering a single data sample), where the marginal distribution at $x_t$ follows a Gaussian whose variance is scaled by some temporal factors. In contrast, under the oracle setting, the marginal distribution at $x_t$ is a mixture of $N$ Gaussians, where $N$ denotes the number of samples. Considering the expression of $\gamma_i(x_t, t)$ (Eq.~\ref{eq:gamma_closed}) in a $D$-dimensional space, as the squared distances scale with $D$ and enter the exponent divided by $2\sigma_t^2$, even modest differences in distance translate into exponentially large differences in weight as $D$ grows (for fixed $\sigma_t$). As a result, once $\sigma_t$ becomes small compared to the typical inter-sample distance, the posterior $\gamma_i(x_t, t)$ becomes sharply peaked on the nearest sample.

\vspace{-8pt}
We verify our claim in Fig.~\ref{fig:split_point}, where the top-1 posterior weight is plotted under varying dimensionalities and sample sizes. Overall, increasing the data dimension leads to faster saturation, whereas increasing the sample size mitigates this effect. 
For ImageNet at $256^2$ resolution, we typically have latent $D \in \{4096, 8192\}$, $N \approx 1400$, yielding the observed top-1 posterior saturation around $t=0.1$. This implies that \textit{the effective training target naturally varies across datasets, even under the same rectified flow objective}. We also note that our demonstration adopts synthetic unit-Gaussian samples; while real data distributions may deviate slightly, their behavior remains similar as in Fig.~\ref{fig:two_stage}. The influence of latent space structure is further analyzed in Sec.~\ref{subsec:latent_space}.

\vspace{4pt}
\begin{findingbox}
\textbf{Takeaways:}
\begin{itemize}[leftmargin=12pt, itemsep=1pt, parsep=0pt, topsep=2pt]
    \item The oracle velocity field reveals a \textbf{two-stage}
    training target: a \textcolor{generalization}{navigation} stage guided by a mixture
    of data modes, followed by a \textcolor{memorization}{refinement} stage dominated
    by the nearest data sample.
    \item The stage transition is jointly governed by the data dimensionality $D$, the sample size $N$, the noise coefficient $\sigma_t$, and the latent space structure.
\end{itemize}
\end{findingbox}

%% file: sec/3_experiment.tex
\section{Model Behaviors under Two-Stage Target}
\label{sec:model_behavior}

\begin{figure*}[t!]
\centering
\includegraphics[width=0.98\linewidth]{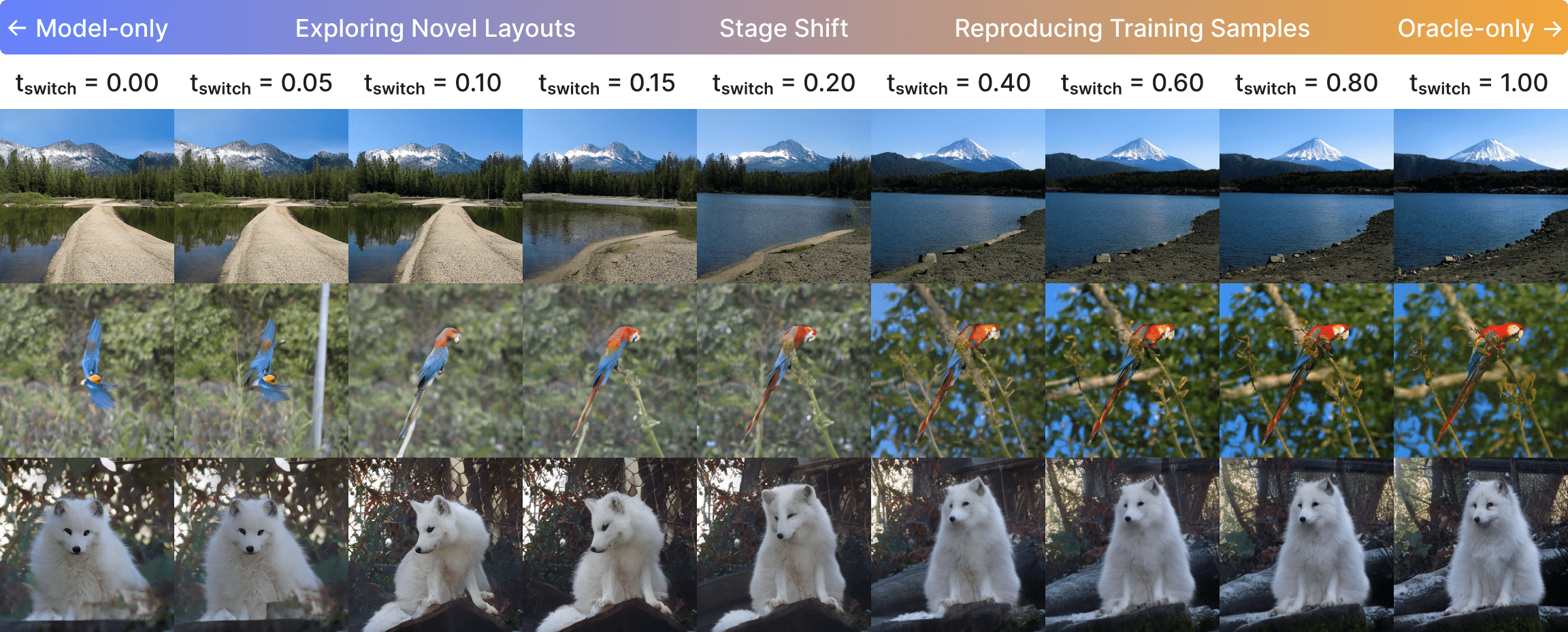}
\vspace{-4pt}
\caption{Mixed sampling results with switch point $t_{\text{switch}}$. Oracle $u_t^*$ is used before $t_{\text{switch}}$ and LightningDiT-XL/1~\cite{yao2025reconstruction} afterward. Overall, early switching yields diverse novel outputs (\textcolor{generalization}{generalization}), while late switching reproduces training samples (\textcolor{memorization}{memorization}).}
\label{fig:qual_mix_sampling}
\vspace{-10pt}
\end{figure*}

\subsection{Observing Empirical Stage Transitions}

We next examine whether a trained flow-based model exhibits analogous stage-specific behaviors characterized by the two-stage target. Specifically, we adopt a LightningDiT-XL/1~\cite{yao2025reconstruction} model and visualize its intermediate predictions, which are calculated by taking a single Euler step from an intermediate timestep $t'\in[0, 1]$ to $t=1$ based on the velocity prediction at $x_{t'}$. 
While diffusion models are designed to progressively denoise noisy inputs, our study aims to identify the transition in model behavior by analyzing the temporal evolution of intermediate predictions.
As demonstrated in Fig.~\ref{fig:qual_model_pred}, several observations arise. First, the model’s prediction at the prior $t=0$ collapses to a coarse class-mean (e.g., predominantly blue for shark, black-and-white for panda). As the trajectory progresses, the early stage primarily navigates the global image layout, which stabilizes around $t\approx0.2$. The later stage focuses exclusively on refining local visual details with minimal semantic or structural deviation, which complies with the consistent training target in this stage. In addition, we also find that \textit{the stage transition in model behavior occurs slightly later than the stage split implied by the oracle target}. One potential explanation for this is that the model may require additional temporal margin to rectify its accumulated prediction errors after shifting to the consistent, CFM-like training target.

\subsection{Relations to Memorization/Generalization}

Moving forward, we further interpret the model's \textcolor{generalization}{generalization} (i.e., the ability to generate novel samples) and \textcolor{memorization}{memorization} (i.e., the tendency to reproduce training samples) behaviors through the lens of stage-level insights. We first consider a mixed sampling scheme: starting from a random Gaussian prior, we take Euler steps based on the oracle velocity $u_t^*$ until a threshold $t_{\text{switch}}$, then switch to the model's velocity predictions. Notably, using the oracle in the first stage ensures that the intermediate states stay on the interpolated distribution, isolating imperfect model predictions. As shown in Fig.~\ref{fig:qual_mix_sampling}, when the oracle is applied throughout the entire trajectory, the process deterministically retrieves a training sample; when $t_{\text{switch}} \in (0.2, 1.0]$, the model can largely replicate the training trajectory, producing images that closely resemble training images. We attribute such \textcolor{memorization}{memorization} behaviors to the trivial and consistent training target in this regime. In contrast, when $t_{\text{switch}} \in [0.0, 0.2]$, the strong prior corruption prevents the model from inferring the original training trajectory, thus deviating from the training instance and exhibiting \textcolor{generalization}{generalization} capability. These results also demonstrate that memorization-from-scratch (i.e., sampling from the prior and recovering exact training samples) is highly unlikely when the dataset size substantially exceeds the effective memorization capacity of the model; in contrast, resuming from training trajectories in the refinement stage is very likely to trigger memorization behaviors. Similar observations have also been reported on smaller-scale datasets~\cite{bertrand2025on}.

\begin{figure}[t!]
\centering
\includegraphics[width=0.98\linewidth]{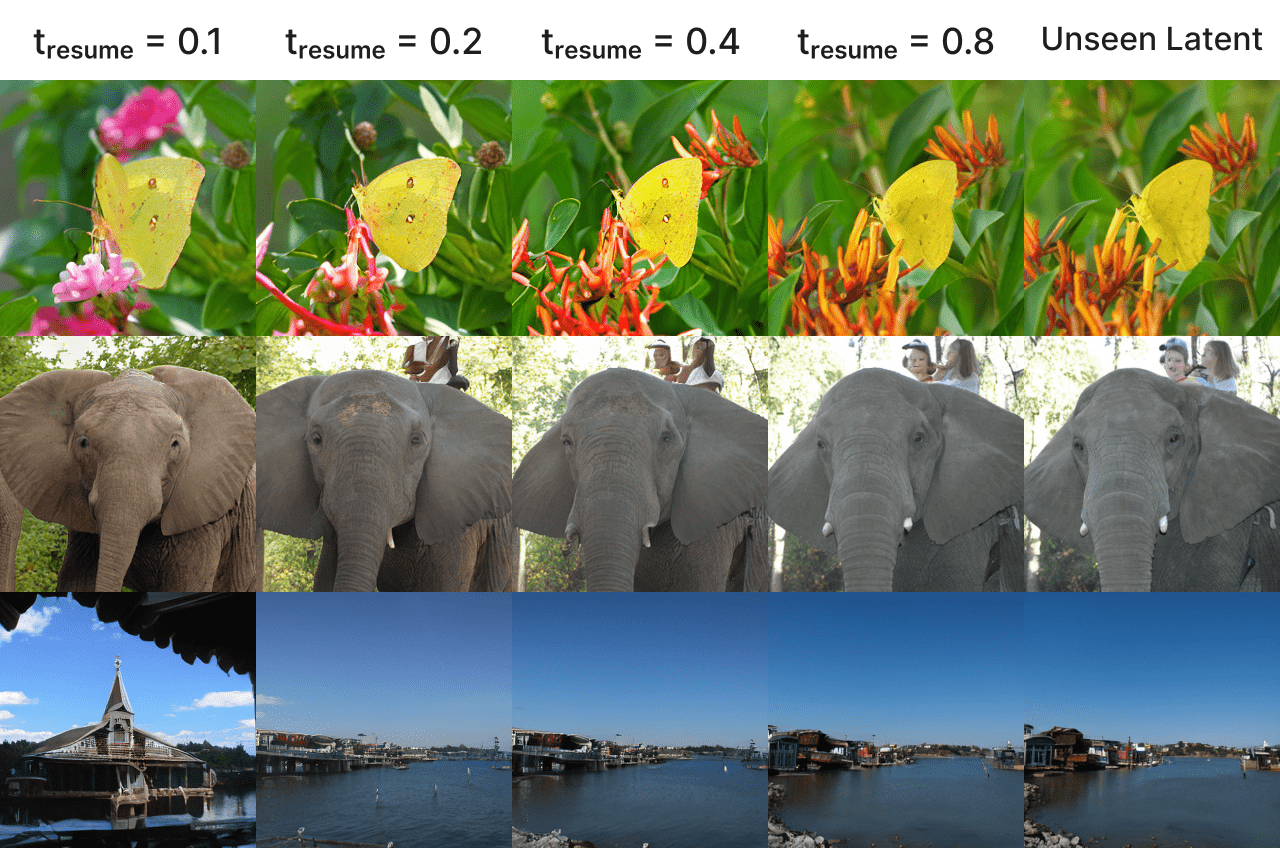}
\vspace{-4pt}
\caption{Qualitative results for refinement generalization. When resuming from $t_\text{resume} \geq 0.2$ on validation image latents, the model largely preserves global structure while improvising fine details.}
\label{fig:denoise_generalization}
\vspace{-8pt}
\end{figure}

Beyond training trajectories, we investigate whether the model’s refinement capability generalizes to unseen data. We construct $x_t$ by combining Gaussian priors and image latents from the ImageNet validation split, resuming the sampling process at $t_{\text{resume}}$. As illustrated in Fig.~\ref{fig:denoise_generalization}, for $t_{\text{resume}}\geq0.2$ the global layout is maintained, while still introducing slight variations in fine details. We also examine the two-stage behaviors on Flux.1~\cite{flux2023} in the Appendix. \looseness=-1

\begin{findingbox}
\textbf{Takeaway:} \textcolor{generalization}{Generalization} stems from the \textcolor{generalization}{navigation} stage, where recovering training trajectories is difficult; \textcolor{memorization}{memorization} arises in the \textcolor{memorization}{refinement} stage, where targets are trivial and consistent.
\end{findingbox}

\subsection{Learning Differs in the Two Stages}
\label{subsec:learning_differs}

\begin{figure}[t!]
\centering
\includegraphics[width=0.98\linewidth]{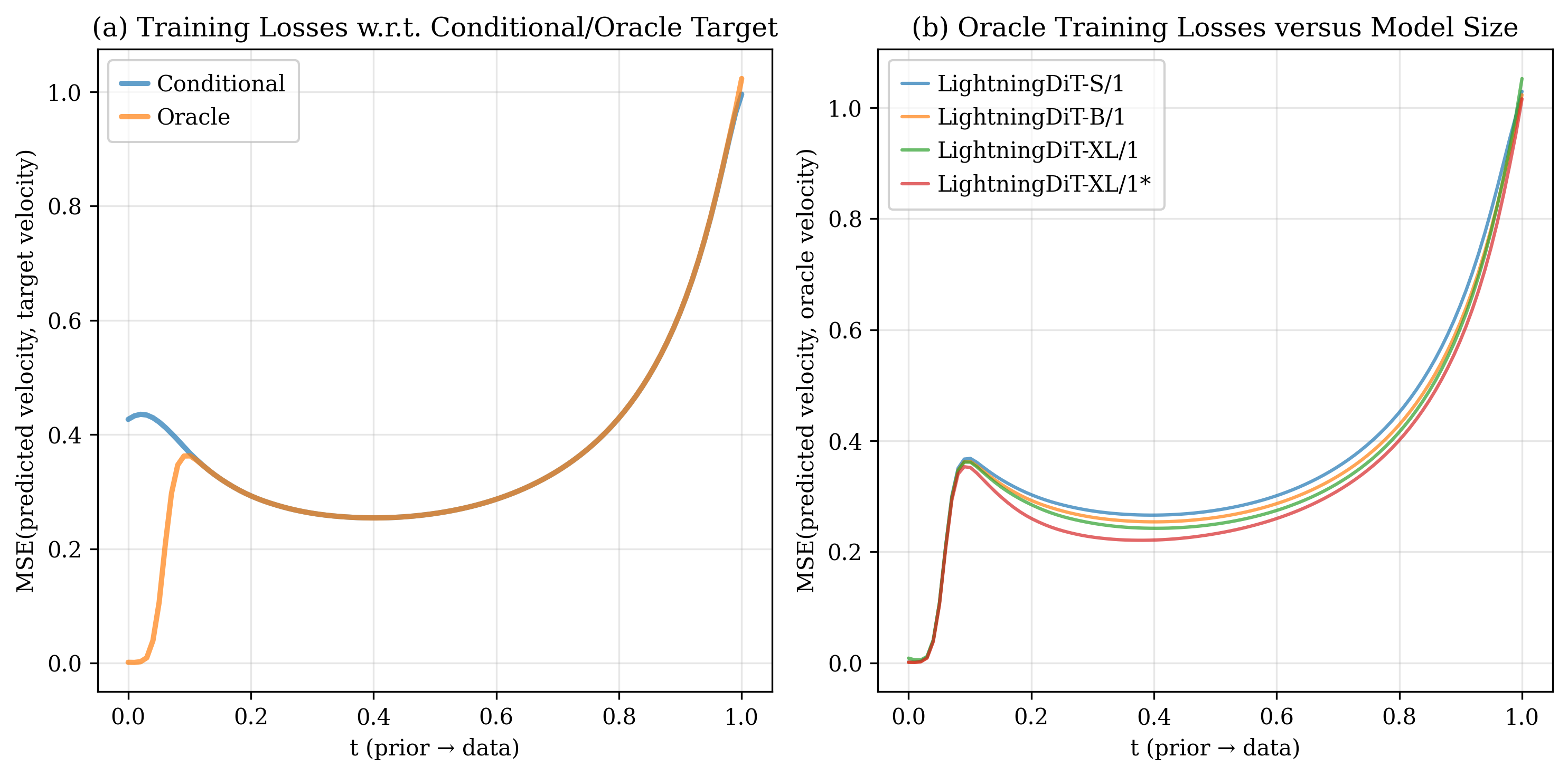}
\vspace{-4pt}
\caption{Training loss trends across timesteps. (a) Training losses (MSE) with respect to conditional and oracle targets. (b) Oracle training losses under varying model sizes. Most models are trained for 100 epochs on the ImageNet~\cite{deng2009imagenet} dataset. * indicates the model is trained for 800 epochs ($8\times$ training compute).}
\label{fig:training_loss}
\vspace{-8pt}
\end{figure}

Our goal is to assess how the model’s learning difficulty varies across timesteps and whether it aligns with the two-stage structure. As shown in Fig.~\ref{fig:training_loss}(a), we plot the training MSE across timesteps by comparing its predicted velocity (without classifier-free guidance~\cite{ho2022cfg}) to both the noisy conditional target $(x_1 - x_0)$ and the class-conditioned oracle target $u_t^*(x_t, t \mid y)$. Consistent with the discrepancy between these targets, \textit{the loss divergence is concentrated in the \textcolor{generalization}{navigation} stage}, while both curves align closely in the refinement stage where their effective forms coincide. Notably, the model achieves near-perfect oracle loss when $t\to 0$, consistent with the trivial class-mean prediction under extreme noise in Fig.~\ref{fig:qual_model_pred}. The oracle loss then rises steadily through the navigation stage, reflecting the increasing challenge of steering toward specific data modes. At the transition into the refinement stage, we observe a noticeable bump in loss, followed by a dip in the mid-refinement region where the intermediate predictions are typically smoothed images. The loss climbs again as $t\to 1$, possibly due to the diversity of potential fine-detail refinements near the data manifold. Importantly, while sharing similar overall trends, \textit{the oracle loss plot is inherently dependent on the latent space structures}; we provide further comparison between VA-VAE~\cite{yao2025reconstruction} and SD-VAE~\cite{rombach2022ldm} in Fig.~\ref{fig:latent_loss}.

Fig.~\ref{fig:training_loss}(b) investigates how model capacity and training compute influence the oracle loss across timesteps. Interestingly, \textit{all models, regardless of parameter size or training duration, exhibit nearly identical performance in the early \textcolor{generalization}{navigation} stage}. In contrast, \textit{the divergence emerges in the vast \textcolor{memorization}{refinement} stage: larger models or those trained with substantially more compute achieve noticeably lower oracle losses}. These results suggest that additional capacity and optimization primarily benefit the refinement of high-frequency details. In addition, improvements to the navigation stage remain limited, possibly due to the difficulty of inferring global structure from near-prior noise.

\begin{table}[t!]
  \centering
  \scriptsize
  \begin{tabular}{cccc}
    \toprule
    \textbf{Sampling Interval} & \textbf{Stage 1 Model} & \textbf{Stage 2 Model} & \textbf{gFID@50K $\downarrow$} \\
    \midrule
    \multirow{4}{*}{$[0.0, 0.1] + [0.1, 1.0]$}
      & XL   & XL   & 2.94 \\
      & Base   & XL   & 3.71 \\
      & XL & Base & 11.26 \\
      & Base & Base & 12.45 \\
    \midrule
    \multirow{4}{*}{$[0.0, 0.2] + [0.2, 1.0]$}
      & XL   & XL   & 2.60 \\
      & Base   & XL   & 4.47 \\
      & XL & Base & 9.24 \\
      & Base & Base & 12.01 \\
    \bottomrule
  \end{tabular}
  \vspace{-4pt}
  \caption{gFID performance under different model combinations. We adopt the Base and XL variants of LightningDiT~\cite{yao2025reconstruction}. ``Stage 1 Model" is used on the first sampling interval, while ``Stage 2 Model" is used on the second. We evenly assign 25 uniform sampling steps to each sub-interval for fair comparison (no CFG).}
  \label{tab:model_mix}
  \vspace{-4pt}
\end{table}

To further validate if the navigation performance is insensitive to model capacity, we conduct a mixed-generation experiment (Tab.~\ref{tab:model_mix}). In this setup, we explicitly swap different models for Stage 1 and Stage 2 during sampling while keeping the total NFE quota (i.e., number of function evaluations in sampling) fixed. Consistent with the oracle-loss trends, replacing the Stage 1 model with a smaller-capacity variant yields little degradation in gFID, indicating that \textit{coarse layout prediction under near-prior noise is largely unaffected by model size}. In contrast, substituting the Stage 2 model with a weaker one leads to a substantial drop in generation quality, confirming that additional capacity primarily benefits the refinement stage.

\begin{findingbox}
\textbf{Takeaway:} \textcolor{generalization}{Navigation} performance is largely invariant to model capacity, whereas \textcolor{memorization}{refinement} benefits significantly from larger models and longer training.
\end{findingbox}

\section{Elucidating Stage-aware Practices}
\label{sec:implications}

Building on the two-stage perspective developed in earlier sections, we now revisit several widely used empirical practices through stage-related insights. Our goal is to clarify why these techniques work and how their effects can be better understood and optimized by explicitly considering the distinct roles of navigation and refinement.

\begin{table}[t!]
  \centering
  \footnotesize
  \begin{tabular}{lcc}
    \toprule
    \textbf{Timestep Shift} & \textbf{Percentage of $t \in [0.0, 0.2]$} & \textbf{gFID@50K $\downarrow$} \\
    \midrule
    $s = 4.0$ & 6\% & 18.89 \\
    $s = 2.0$ & 12\% & 14.82 \\
    \midrule
    $s = 1.0$ (uniform) & 22\% & 12.99 \\
    \midrule
    $s = 0.7$ & 28\% & 12.46 \\
    $s = 0.5$ & 34\% & \textbf{12.23} \\
    $s = 0.3$ & 46\% & 12.66 \\
    $s = 0.1$ & 72\% & 19.91 \\
    \bottomrule
  \end{tabular}
  \vspace{-4pt}
  \caption{Generation performance with different timestep shift factors (NFE = 50, no CFG). We adopt a LightningDiT-B/1~\cite{yao2025reconstruction} model for evaluation. Overall, a moderate increase in early (\textcolor{generalization}{navigation}) steps gives the best trade-off in sampling quality.}
  \label{tab:shift_factor}
  \vspace{-4pt}
\end{table}

\begin{table}[t!]
  \centering
  \footnotesize
  \setlength{\tabcolsep}{5pt}
  \begin{tabular}{cc|cc}
    \toprule
    \textbf{CFG Interval} & \textbf{gFID@50K $\downarrow$} &
    \textbf{CFG Interval} & \textbf{gFID@50K $\downarrow$} \\
    \midrule
    None & 12.99 & $[0.0,\,1.0]$ & 10.79 \\
    \midrule
    $[0.0, 0.1]$ & 6.33 & $[0.0, 0.2]$ & 6.62 \\
    $[0.1, 0.2]$ & \textbf{5.21} & $[0.0, 0.4]$ & 8.03 \\
    $[0.2, 0.3]$ & 7.70 & $[0.0, 0.6]$ & 9.19 \\
    $[0.3, 0.4]$ & 9.39 & $[0.0, 0.8]$ & 10.14 \\
    $[0.4, 0.5]$ & 10.39 & $[0.1, 0.3]$ & 3.54 \\
    $[0.5, 0.6]$ & 10.59 & $[0.1, 0.4]$ & 4.16 \\
    $[0.6, 0.7]$ & 11.06 & $[0.1, 0.5]$ & 2.82 \\
    $[0.7, 0.8]$ & 11.38 & $[0.1, 0.6]$ & \textbf{2.80} \\
    $[0.8, 0.9]$ & 11.64 & $[0.1, 0.7]$ & 2.86 \\
    $[0.9, 1.0]$ & 12.20 & $[0.1, 0.8]$ & 2.97 \\
    \bottomrule
  \end{tabular}
  \vspace{-4pt}
  \caption{Generation performance under different CFG intervals (CFG factor $\omega$ = 2.5). We adopt a LightningDiT-B/1~\cite{yao2025reconstruction} model for evaluation. Optimal intervals concentrate in the early-mid \textcolor{memorization}{refinement} stage, while excessive guidance in the earliest \textcolor{generalization}{navigation} steps can degrade the sampling quality.}
  \label{tab:cfg_interval}
  \vspace{-8pt}
\end{table}

\subsection{Optimizing Timestep Schedule}

A common practice in diffusion sampling is to use a uniform timestep schedule over $t \in [0, 1]$. However, a natural question arises: \emph{what is the optimal allocation of computation between the navigation and refinement stages under a fixed NFE budget?} We narrow down our discussion to \textit{timestep shifting}~\cite{esser2024sd3}, an existing technique for constructing non-uniform timestep schedules. Specifically, timestep shifting transforms a uniform timestep $t_n$ with a smooth monotonic mapping: $t_m = \frac{s \, t_n}{1 + (s - 1)t_n}$, where $s$ is a shift factor. With $s<1$, we allocate more steps to the early (navigation) timesteps, while $s>1$ biases the schedule toward later (refinement) timesteps. While this technique was originally proposed to address noise-level imbalance in high-resolution image generation, this approach naturally provides a flexible knob to control stage-wise sampling step allocation. The results in Tab.~\ref{tab:shift_factor} indicate that biasing the timestep distribution slightly toward earlier steps leads to noticeably better samples, suggesting that \textit{modestly emphasizing the \textcolor{generalization}{navigation} phase is beneficial for inference}.

\subsection{Optimizing CFG Intervals}

Classifier-free guidance (CFG)~\cite{ho2022cfg} modulates the conditional signal during sampling by amplifying the difference between conditional and unconditional predictions. Prior work has shown that applying CFG only on a selected sub-interval, rather than across all timesteps, leads to improved generation quality~\cite{kynkaanniemi2024applying}. We revisit this idea in the context of flow-based diffusion models with results in Tab.~\ref{tab:cfg_interval}. Our ablations reveal three consistent trends. First, \textit{the most effective single short interval (of width 0.1) lies in the transition stage between \textcolor{generalization}{navigation} and \textcolor{memorization}{refinement}}. Second, when the CFG interval is expanded, it becomes beneficial to exclude the very initial segment $[0.0,0.1]$, likely because amplified guidance at extremely noisy states interferes with the formation of stable global layouts. Third, \textit{the overall optimal CFG ranges tend to span the early and mid \textcolor{memorization}{refinement} stage}. This coincides with the timesteps where velocity prediction norms peak (see Appendix), suggesting that CFG is most effective during high-confidence refinement.

\subsection{Influence of Latent Space Structure}
\label{subsec:latent_space}

\begin{figure}[t!]
\centering
\includegraphics[width=0.98\linewidth]{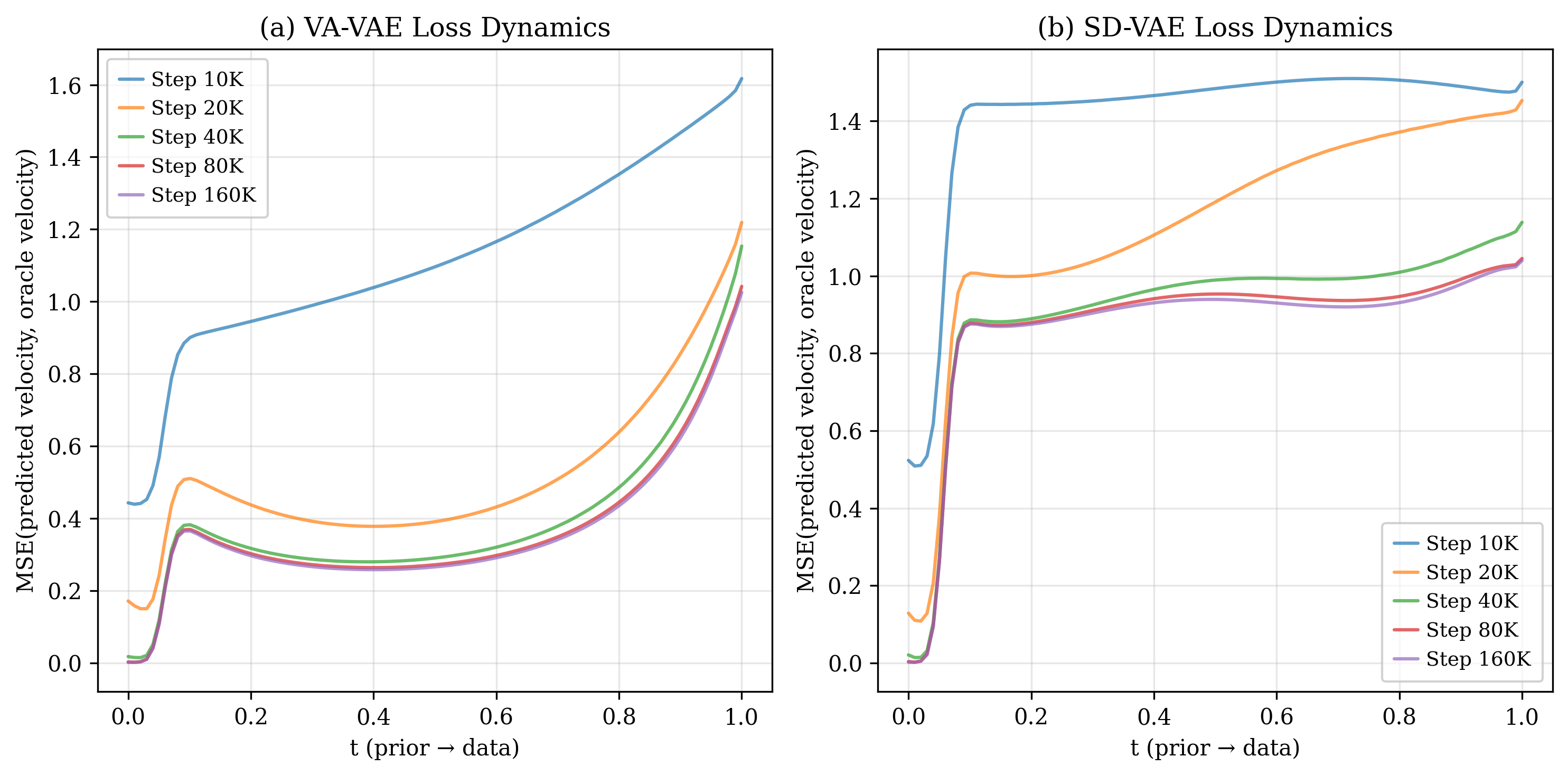}
\vspace{-4pt}
\caption{Convergence of oracle loss under different latent spaces: VA-VAE~\cite{yao2025reconstruction} losses decrease smoothly in the refinement stage; whereas SD-VAE~\cite{rombach2022ldm} exhibits wavy patterns. The gFID dynamics are attached in the Appendix, where VA-VAE converges faster.}
\label{fig:latent_loss}
\vspace{-8pt}
\end{figure}

We next examine how the choice of latent space affects the model’s dynamics under a two-stage oracle target. Fig.~\ref{fig:latent_loss} compares the oracle loss dynamics under VA-VAE~\cite{yao2025reconstruction} and SD-VAE~\cite{rombach2022ldm} latent spaces during training. In the navigation stage, both spaces exhibit a gradual oracle loss increase, suggesting that \textit{the growing difficulty of coarse-layout prediction is agnostic to the latent space}. In the refinement stage, however, the loss trends diverge substantially. While the absolute magnitudes are not directly comparable, certain patterns arise from the shapes of the oracle loss curves. For VA-VAE, whose latent space is aligned with semantic structure through DINO-based VF loss~\cite{oquab2023dinov2, yao2025reconstruction}, the oracle loss converges smoothly toward a parabola-like shape as training progresses. This suggests a well-organized latent space, where semantic modes are arranged coherently. In comparison, SD-VAE is trained purely for low-level reconstruction, so it converges to a flatter curve with noticeable waviness, suggesting that its latent space is less structured and the model struggles to perform consistent denoising across stages. Overall, \textit{latent spaces with high-level concept alignment tend to induce clearer mode organization, facilitating both \textcolor{generalization}{navigation} and \textcolor{memorization}{refinement}}.

\begin{findingbox}
\textbf{Takeaways:}
\begin{itemize}[leftmargin=12pt, itemsep=1pt, parsep=0pt, topsep=2pt]
    \item \textbf{Inference timestep schedule:} Allocating modestly more sampling steps to the \textcolor{generalization}{navigation} stage improves gFID under a fixed NFE budget.
    \item \textbf{CFG interval:} The optimal guidance interval concentrates in the early-to-mid \textcolor{memorization}{refinement} stage.
    \item \textbf{Latent space:} Latent spaces with clearer mode organization yield smoother oracle loss convergence and better generation quality.
\end{itemize}
\end{findingbox}

%% file: sec/4_related_work.tex
\section{Related Work}
\label{sec:related_work}

\subsection{Diffusion-based Generative Models}

Diffusion models learn data distributions through scheduled interpolations to a simple prior (typically Gaussian). This process is governed by the probability flow ODE (PF-ODE)~\cite{song2020score} and can be interpreted through several complementary viewpoints~\cite{lai2025principles}. From a probabilistic perspective, DDPM-style methods define a discrete-time forward noising process and learn to invert it via noise/data prediction targets~\cite{ho2020denoising, dhariwal2021diffusion}. The score matching perspective instead characterizes the dynamics in continuous time using ODE/SDE formulations~\cite{song2019generative, song2020score}. Another line of work sources from continuous normalizing flows (CNFs)~\cite{lipman2023flow} and stochastic interpolants~\cite{albergo2023interpolant, albergo2023stochastic} and unifies to flow matching (FM), where the model directly learns the velocity field governing the transport between prior and data. Recent work favors the rectified flow formulation~\cite{liu2023flow} for its simplicity and stable optimization; whereas emerging research explores new generative paradigms for few-step diffusion, such as ShortCut models~\cite{frans2025shortcut}, MeanFlow~\cite{geng2025mean}, and AlphaFlow~\cite{zhang2025alphaflow}.

\subsection{Memorization versus Generalization}

Understanding the balance between memorization and generalization is a central theme in generative modeling. One line of work focuses on measuring model behaviors through quantitative metrics, such as the effective model memorization (EMM) factor~\cite{gu2024on} and the probability flow distance (PFD)~\cite{zhang2025understanding}. Another line of work investigates the factors influencing the memorization-generalization trade-off: Zhang \textit{et al.}~\cite{zhang2023emergence} observe distinctive generalization and memorization regimes during training; Bonnaire \textit{et al.}~\cite{bonnaire2025why} show that diffusion models possess separate generalization and memorization timescales, with larger datasets widening the effective generalization window during training; Shi \textit{et al.}~\cite{shi2025a} identify that declining entropy in recursively generated training data triggers a memorization-dominated collapse in diffusion models, which can be alleviated through an entropy-based selection strategy; Niedoba \textit{et al.}~\cite{niedoba2025towards} reveal that network denoisers generalize through localized denoising operations; Kadkhodaie \textit{et al.}~\cite{kadkhodaie2024generalization} find that two independently trained denoisers converge to nearly identical score functions.
However, most existing studies focus on small-scale datasets and sampling-from-scratch analysis, limiting their applicability to modern large-scale diffusion models. Song et al.~\cite{song2025selective} observe this trade-off at ImageNet scale, yet without probing the underlying causes. Building on prior explorations of closed-form targets under various diffusion formulations~\cite{bertrand2025on, kamb2025an, biroli2024dynamical, gao2024flow, li2024good, scarvelis2025closedform}, we derive the oracle velocity for rectified flow to dissect the two-stage behaviors between navigation and refinement.

%% file: sec/5_conclusion.tex
\section{Discussion}
\label{sec:discussion}

\subsection{Summary of Best Practices}

The best practices involve three major aspects: \textit{data manifold structure}, \textit{training-time techniques}, and \textit{inference-time techniques}. Firstly, a structured latent space makes generative modeling easier (e.g., by aligning with DINO features). At training time, the closed-form oracle allows us to identify a theoretical stage-shift point, where more training steps can be allocated to the refinement stage for faster convergence (see Appendix). At inference time, generation quality can be further improved by adopting a timestep-shifted sampling schedule (i.e., slightly more navigation steps) and by applying CFG primarily within the refinement stage.

\subsection{Improving Navigation Capability}
Our findings indicate that the navigation performance under the FM/CFM objective remains largely unchanged after scaling model capacity and training compute. This reveals an inherent limitation in the near-prior regime: either the prediction task itself is trivial and all models saturate, or the per-sample CFM target provides insufficient supervision for larger models to benefit. Moreover, standard metrics like generation FID~\cite{heusel2017fid} provide little insight into intermediate prediction quality, underscoring the need for new measures that better capture layout fidelity in the early stage.

\subsection{The Hidden Mechanism behind Scaling}
As the stage transition depends on data dimensionality and dataset size, scaling along these axes directly reshapes the underlying navigation-refinement dynamics: higher dimensionality sharpens the stage transition, while larger datasets broaden the modes the model must navigate and refine. In practice, state-of-the-art diffusion models scale data, model capacity, and training compute simultaneously; the findings from the oracle velocity offer intuition for why scaling systematically stabilizes navigation, enriches denoising, and ultimately improves generative quality.

\subsection{Oracle Velocity for Training}
A natural question is whether the closed-form oracle can directly serve as a training target. Our experiments show that oracle-supervised training is feasible and yields convergence behavior similar to CFM. However, this approach remains impractical at scale due to the computational and I/O overhead of evaluating the oracle over the full dataset. Nonetheless, the closed-form oracle remains a valuable analytical tool for analyzing model behaviors.

\section{Conclusion}
\label{sec:conclusion}

In this work, we introduce a principled framework for understanding diffusion models. By deriving the closed-form oracle velocity, we show that the effective training target is inherently two-stage: a multi-sample guided navigation regime near the prior and a single-sample dominated refinement regime near the data. This perspective explains the divergence between memorization and generalization, clarifies what diffusion models learn and how they sample, and sheds light on widely adopted empirical practices. Our findings provide an oracle-driven viewpoint for advancing the development of next-generation diffusion models.

%% file: sec/X_suppl.tex
\clearpage
\renewcommand{\thesection}{\Alph{section}}
\setcounter{section}{0}

\section*{Appendix Outline}
\noindent The appendix is organized as follows:
\begin{itemize}
    \item Appendix~\ref{sec:supp_proof} provides the proof of Theorem~2.1.
    \item Appendix~\ref{sec:supp_analysis} presents supplementary analysis, including model prediction trends (\ref{subsec:model_pred_trend}), latent space comparisons (\ref{subsec:latent_space_impact}), quantification of memorization behaviors (\ref{subsec:quantify_memorization}), and timestep shift illustration (\ref{subsec:timestep_shift_illustration}).
    \item Appendix~\ref{sec:implementation} lists implementation details.
    \item Appendix~\ref{sec:additional_practices} explores additional training-time practices, such as oracle-supervised training (\ref{subsec:oracle_training}) and stage-aware training timestep sampling (\ref{subsec:nonuniform_training}).
    \item Appendix~\ref{sec:flux_findings} extends our findings to Flux.1[dev]~\cite{flux2023}, including timestep-shifted sampling (\ref{subsec:flux_inference}) and analysis of two-stage behaviors in larger diffusion models (\ref{subsec:flux_two_stage}). 
    \item Appendix~\ref{sec:additional_qual} presents additional qualitative results.
\end{itemize}

\section{Proof of Theorem 2.1}
\label{sec:supp_proof}

The Flow Matching (FM) objective (Eq.~\ref{eq:supp_fm}) is given by:
\begin{equation}
    \mathcal{L}_{\text{FM}}(\theta) = \mathbb{E}_{t, \, p_t(x_t)} || v_t(x_t; \theta) - u_t(x_t) ||^2.
    \label{eq:supp_fm}
\end{equation}
The marginal velocity field $u_t(x_t)$ in Eq.~\ref{eq:supp_fm} is generally intractable under unknown probability paths. However, we want to show that it admits a closed-form solution under the following conditions: (i) a Gaussian prior distribution, (ii) a finite dataset $\{x_1^{(i)}\}_{i=1}^N$ approximating $p_{\text{data}}$, and (iii) a linear interpolation path as in rectified flow. Formally, our goal is to obtain a closed-form expression for the following conditional expectation:
\begin{equation}
u_t^*(x_t,t) \coloneqq \mathbb{E}_{x_1 \sim p_{\text{data}}} \!\left[ u_t(x_t \mid x_1) \,\big|\, x_t \right].
\end{equation}
Under linear flows (i.e., probability paths constructed via linear interpolation), we have:
\begin{equation}
x_t = \alpha_t x_1 + \sigma_t x_0 \quad \Longrightarrow \quad x_0 = \frac{x_t-\alpha_t x_1}{\sigma_t}.
\label{eq:supp_interpolation}
\end{equation}
Recall that the conditional velocity in CFM is given by:
\begin{equation}
u_t(x_t \mid x_1) = \dot \alpha_t x_1 + \dot \sigma_t x_0.
\label{eq:supp_conditonal_velocity}
\end{equation}
Substituting Eq.~\ref{eq:supp_interpolation} into Eq.~\ref{eq:supp_conditonal_velocity} writes $u_t(x_t\mid x_1)$ as a function of $x_t$ and $x_1$:
\begin{equation}
  u_t(x_t\mid x_1)\;=\;\Big(\dot\alpha_t-\frac{\alpha_t\dot\sigma_t}{\sigma_t}\Big)\,x_1\;+\;\frac{\dot\sigma_t}{\sigma_t}\,x_t.
\end{equation}
Taking the conditional expectation given $x_t$ yields:
\begin{align}
  u_t^*(x_t,t)\;&\coloneqq\;\mathbb{E}\!\left[u_t(x_t\mid x_1)\mid x_t\right] \\
  &=\; \left ( \dot\alpha_t-\frac{\alpha_t\dot\sigma_t}{\sigma_t} \right ) \,\mathbb{E}[x_1\mid x_t]\;+\; \frac{\dot\sigma_t}{\sigma_t} \,x_t,
\label{eq:supp_oracle}
\end{align}
Given a finite dataset $\{x_1^{(i)}\}_{i=1}^N$, we essentially approximate the true data distribution $p_{\text{data}}(x_1)$ via an empirical mixture:
\begin{equation}
p_{\text{data}}(x_1) \approx \frac{1}{N} \sum_{i=1}^N \delta (x_1 - x_1^{(i)}),
\end{equation}
where $\delta(\cdot)$ is the Dirac delta function  with $\delta(0) = \infty$ and zero elsewhere. Accordingly, the empirical probability path marginal $\tilde p_t(x_t)$ is given by a Gaussian mixture:
\begin{equation}
\tilde p_t(x_t)
  \;=\;\frac{1}{N}\sum_{i=1}^N \mathcal{N}\!\big(x_t;\,\alpha_t x_1^{(i)},\,\sigma_t^2 I\big).
\end{equation}
By Bayes’ rule, the posterior $p(x_1^{(i)}\mid x_t)$ is proportional to $p(x_t\mid x_1^{(i)})$ up to a common normalizing factor that ensures the probabilities sum to one. Since all mixture components share the same Gaussian covariance $\sigma_t^2I$ and uniform prior weight $1/N$, their normalization constants cancel out when computing the posterior weights. Let $\gamma_i(x_t,t)$ denote the resulting normalized weighting function that reflects the relative contribution of each data sample $x_1^{(i)}$ to the current point $x_t$, we have:
\begin{equation}
\gamma_i(x_t,t)\;=\;\frac{\exp\!\big(-\tfrac{\|x_t-\alpha_t x_1^{(i)}\|^2}{2\sigma_t^2}\big)}
  {\sum_{j=1}^N \exp\!\big(-\tfrac{\|x_t-\alpha_t x_1^{(j)}\|^2}{2\sigma_t^2}\big)}.
\label{eq:supp_gamma}
\end{equation}
Hence, the posterior mean is given by:
\begin{equation}
\mathbb{E}[x_1\mid x_t]\;=\;\sum_{i=1}^N \gamma_i(x_t,t)\,x_1^{(i)}.
\label{eq:supp_posterior_mean}
\end{equation}
This is also known as the Nadaraya-Watson estimator~\cite{nadaraya, watson}. Combining Eq.~\ref{eq:supp_oracle} and~\ref{eq:supp_posterior_mean}, we reach the closed form:
\begin{equation}
  u_t^*(x_t,t)
  \;=\;A_t\sum_{i=1}^N \gamma_i(x_t,t)\,x_1^{(i)}\;+\;B_t\,x_t,
\end{equation}
where $A_t=\dot\alpha_t-\frac{\alpha_t\dot\sigma_t}{\sigma_t}$, $
B_t=\frac{\dot\sigma_t}{\sigma_t}$. We refer to this closed-form expression of the marginal velocity field under the linear probability path construction as the \textit{oracle velocity field}. Moreover, the oracle velocity field can also be evaluated conditionally; for instance, under a class-conditional generation setting, it can be computed within each class-specific subset, denoted as $u_t^*(x_t,t \mid y)$. 
\hfill $\square$

\newpage
\section{Supplementary Analysis}
\label{sec:supp_analysis}

\subsection{Model Prediction Trends}
\label{subsec:model_pred_trend}

\begin{figure}[h!]
\centering
\includegraphics[width=0.98\linewidth]{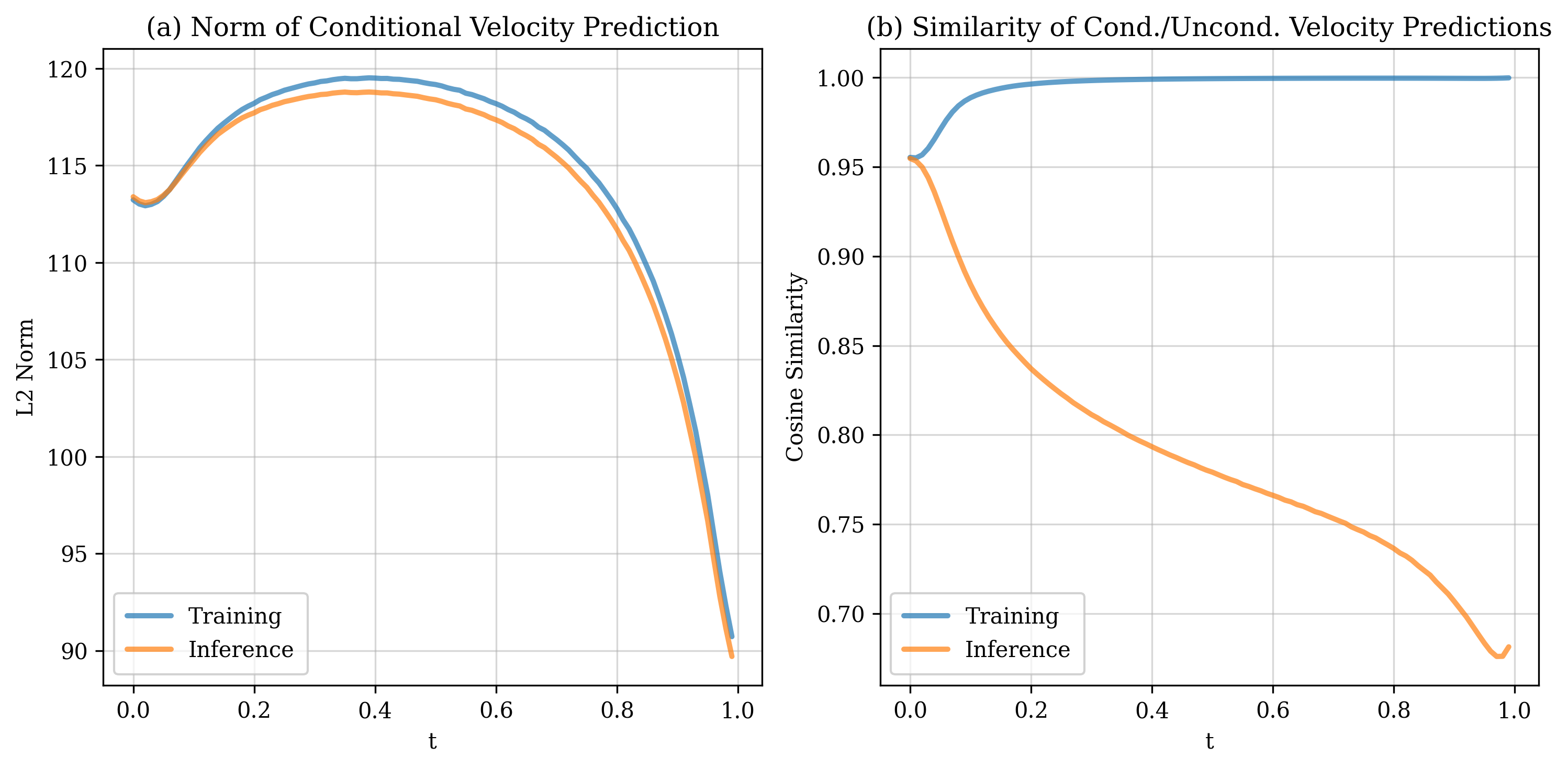}
\vspace{-8pt}
\caption{Analysis of model prediction trends. (a) Norm of velocity predictions peaks around $t{=}0.4$, coinciding
with the timesteps where CFG most effectively enhances
sample fidelity (Tab.~\ref{tab:cfg_interval}). (b) Cosine similarity between conditional and unconditional predictions. Along
training trajectories, the two remain aligned; during inference, $x_t$ deviates from the supervised region, leading to diverged predictions and exhibiting
generalization behaviors.}
\label{fig:model_pred}
\vspace{-8pt}
\end{figure}

\subsection{Latent Space Comparisons}
\label{subsec:latent_space_impact}

\begin{figure}[h!]
\centering
\includegraphics[width=0.85\linewidth]{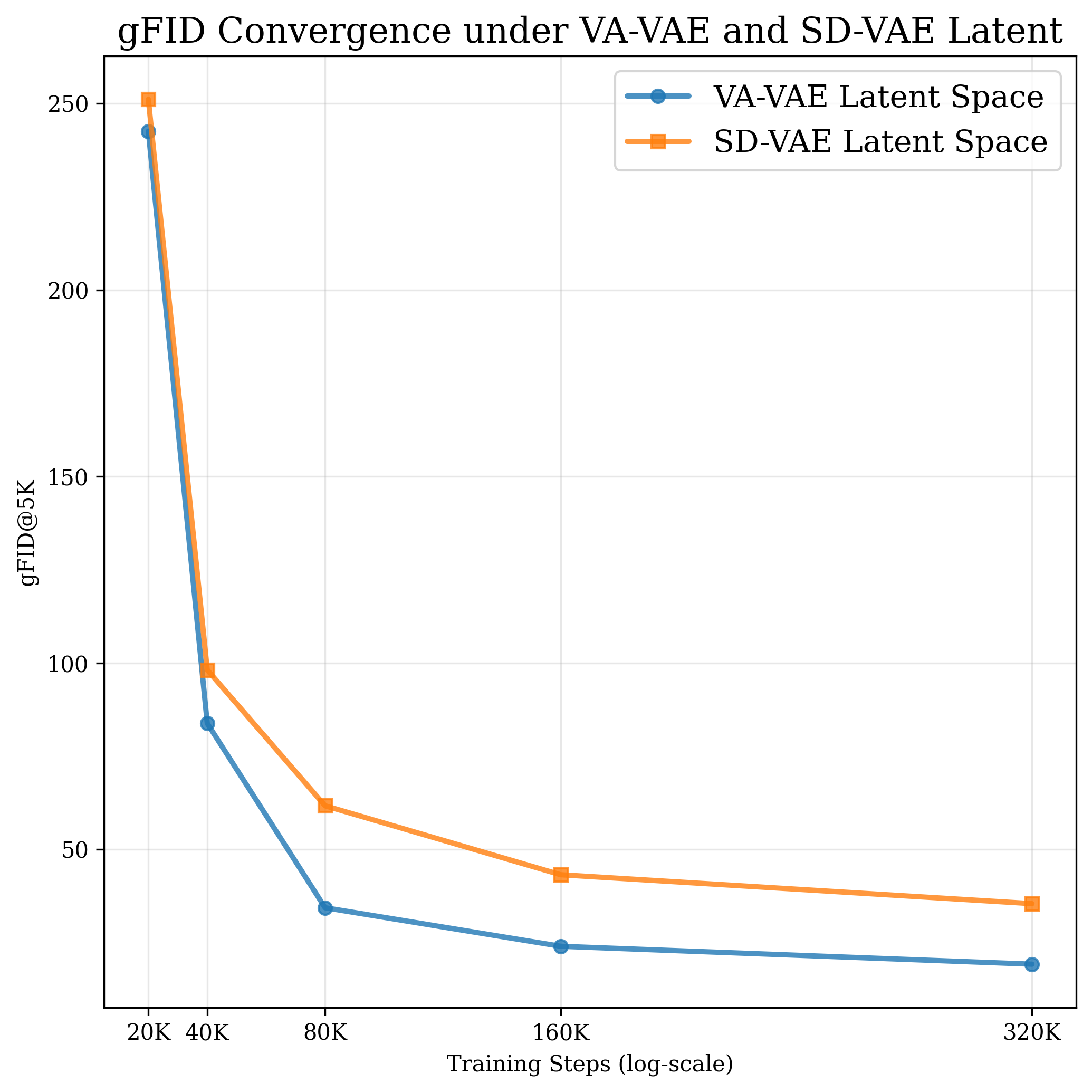}
\vspace{-4pt}
\caption{Convergence of gFID@5K when training rectified flow models under different latent spaces. We use LightningDiT-B/1 for VA-VAE~\cite{yao2025reconstruction} and LightningDiT-B/2 for SD-VAE~\cite{rombach2022ldm} to align the training resolution to $16^2$. The training in the VA-VAE latent space converges faster, indicating a better latent space structure.}
\vspace{-4pt}
\label{fig:supp_latent_gfid}
\end{figure}

\newpage
\subsection{Quantification of Memorization Behaviors}
\label{subsec:quantify_memorization}

\begin{figure}[h!]
\centering
\includegraphics[width=0.8\linewidth]{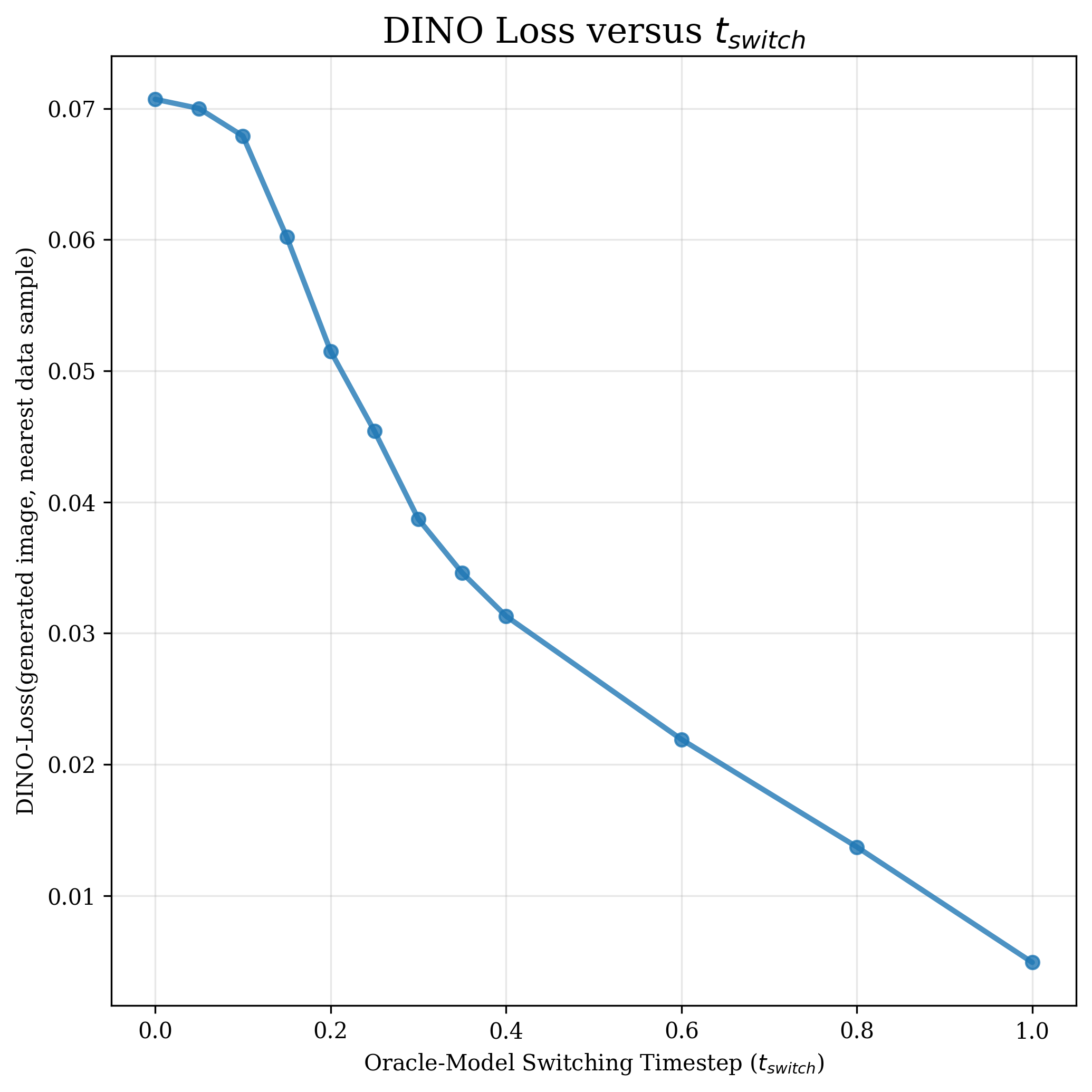}
\vspace{-4pt}
\caption{Quantitative results for oracle-model mixed generation. We report the DINO~\cite{oquab2023dinov2} loss (MSE of DINO self-attention maps, reflecting structural dissimilarity) between the generated images and the nearest training sample, measured across different switching timesteps ($t_{\text{switch}}$). Overall, we observe: (1) a sharp decline emerges after $t_{\text{switch}} \approx 0.1$ (i.e., the shift of training target); (2) when the loss falls below roughly 0.05, the generated layouts become closely aligned with those of the training samples (Fig.~\ref{fig:supp_mix_gen}).}
\vspace{-4pt}
\label{fig:supp_dino_loss}
\end{figure}

\subsection{Timestep Shift Illustration}
\label{subsec:timestep_shift_illustration}

\begin{figure}[h!]
\centering
\includegraphics[width=0.8\linewidth]{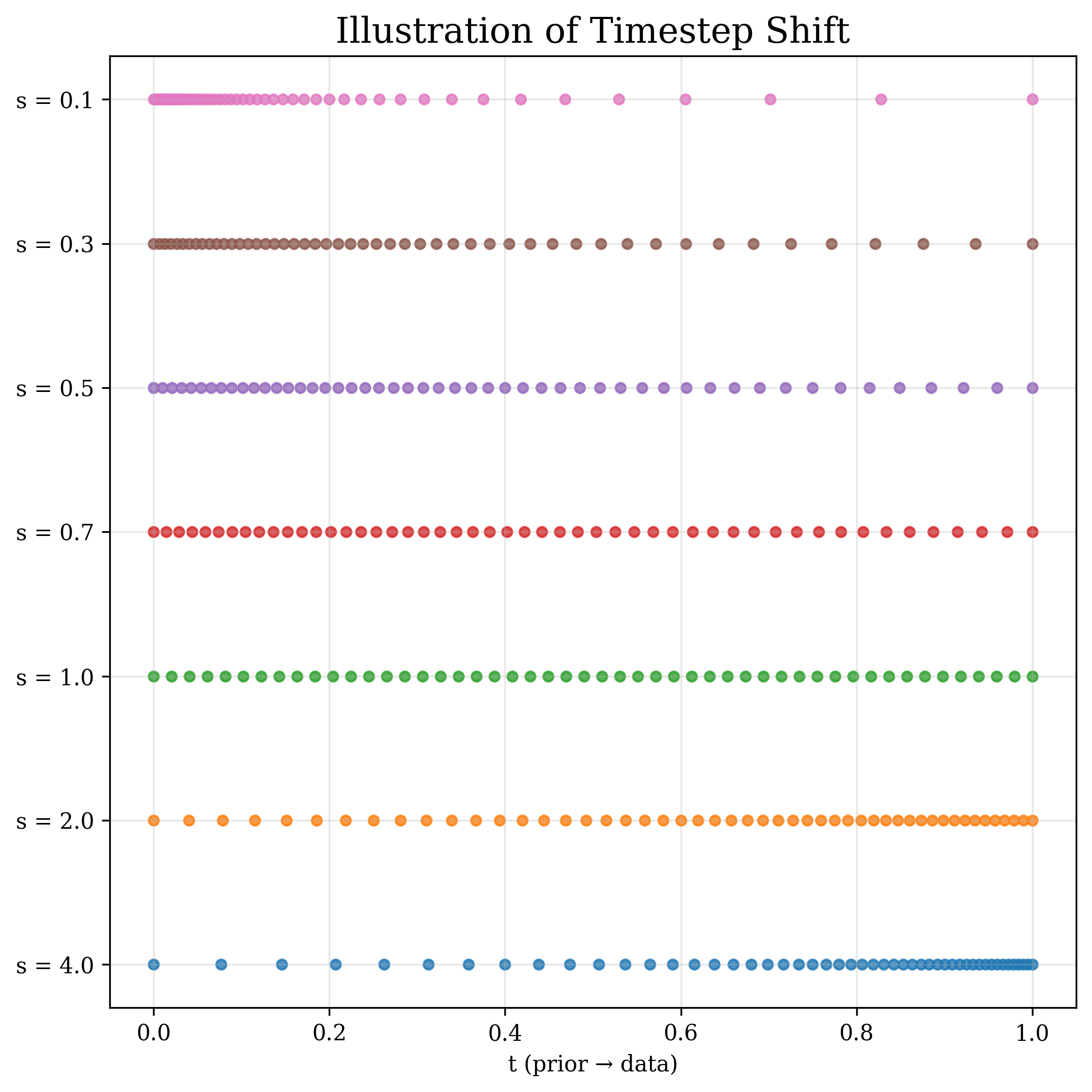}
\vspace{-4pt}
\caption{Illustration of timestep shift mapping $t_m = \frac{s \, t_n}{1 + (s - 1)t_n}$, where $s$ is a shift factor and $t_n$ is the uniform sampling schedule. Intuitively, with $s<1$, we allocate more steps to the early (\textcolor{generalization}{navigation}) timesteps, while $s>1$ biases the schedule toward later (\textcolor{memorization}{refinement}) timesteps. The best gFID is achieved with $s = 0.5$.}
\vspace{-8pt}
\label{fig:supp_timestep_shift}
\end{figure}

\newpage
\section{Implementation Details}
\label{sec:implementation}

\begin{table}[h!]
\centering
\scriptsize
\setlength{\tabcolsep}{6pt}
\renewcommand{\arraystretch}{1.15}
\begin{tabular}{l|ccc}
\toprule
\rowcolor[gray]{0.9}
\multicolumn{4}{l}{\textbf{architecture}} \\
LightningDiT~\cite{yao2025reconstruction} variants & Small & Base & XL \\
depth                     & 12   & 12   & 28   \\
hidden dim                & 384  & 768 & 1152 \\
heads                     & 6   & 12   & 16   \\
image size                & \multicolumn{3}{c}{256} \\
patch size                & \multicolumn{3}{c}{1 (VA-VAE), 2 (SD-VAE)} \\
latent size               & \multicolumn{3}{c}{$16 \times 16$} \\
\midrule
\rowcolor[gray]{0.9}
\multicolumn{4}{l}{\textbf{training}} \\
epochs                    & \multicolumn{3}{c}{$\{100, 800\}$} \\
optimizer                 & \multicolumn{3}{c}{AdamW~\cite{loshchilov2018adamw} ($\beta_1, \beta_2 = 0.9, 0.995$)} \\
batch size                & \multicolumn{3}{c}{512} \\
learning rate             & \multicolumn{3}{c}{1e-4} \\
learning rate schedule    & \multicolumn{3}{c}{constant} \\
weight decay              & \multicolumn{3}{c}{0} \\
max gradient norm         & \multicolumn{3}{c}{1.0} \\
ema decay                 & \multicolumn{3}{c}{0.9999} \\
time sampler              & \multicolumn{3}{c}{Uniform[0, 1]} \\
class token drop (for CFG)   & \multicolumn{3}{c}{0.1} \\
\midrule
\rowcolor[gray]{0.9}
\multicolumn{4}{l}{\textbf{sampling}} \\
ODE solver                & \multicolumn{3}{c}{Euler} \\
ODE steps                 & \multicolumn{3}{c}{50} \\
time steps                & \multicolumn{3}{c}{uniform / stage-wise uniform / shifted} \\
CFG~\cite{ho2022cfg} scale                 & \multicolumn{3}{c}{$\{1.0, 2.5\}$} \\
\bottomrule
\end{tabular}
\caption{Implementation details.}
\label{tab:supp_configurations}
\end{table}

\section{Additional Training-Time Practices}
\label{sec:additional_practices}

\subsection{Oracle-Supervised Training}
\label{subsec:oracle_training}

\begin{figure}[h!]
\centering
\includegraphics[width=0.78\linewidth]{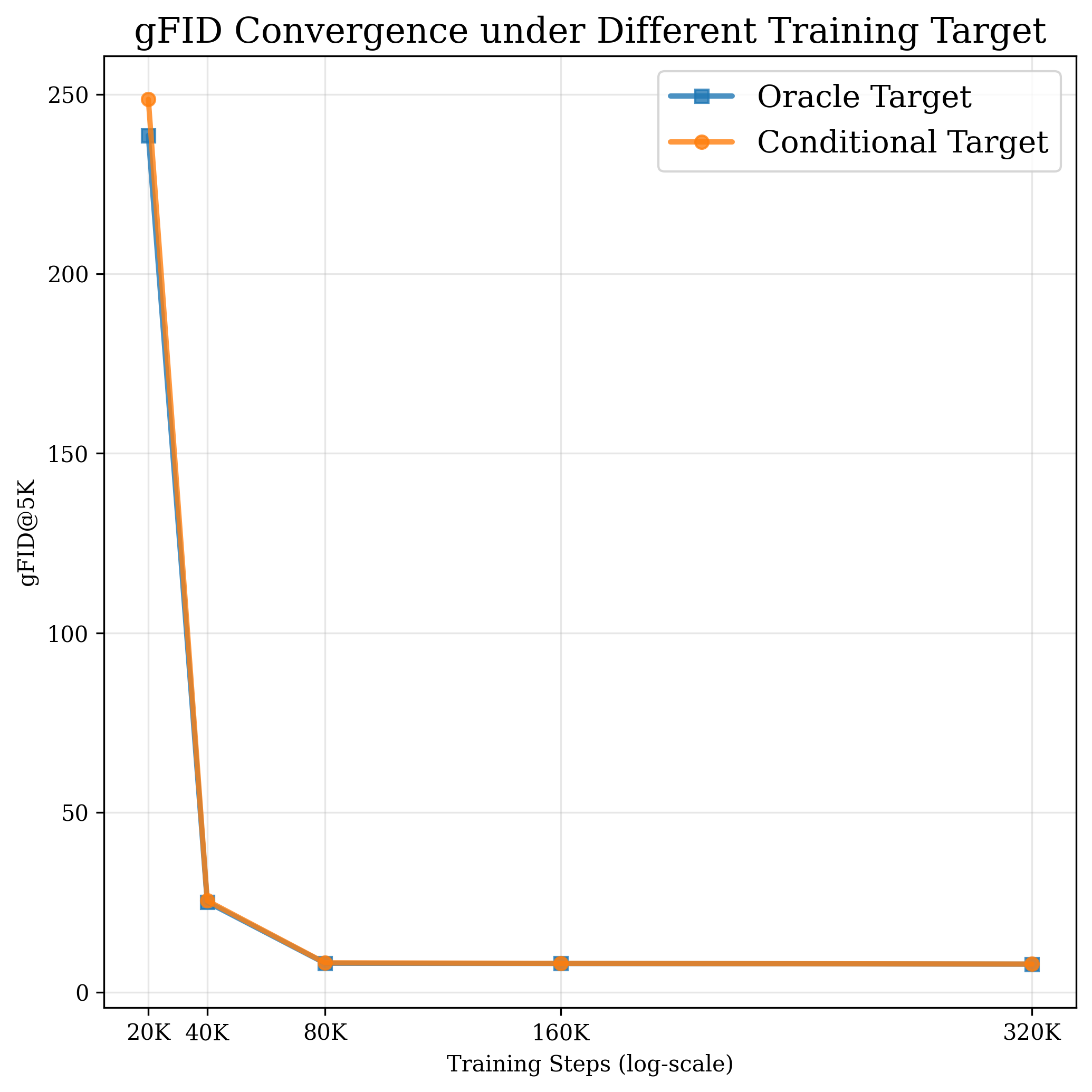}
\vspace{-4pt}
\caption{Convergence of gFID@5K when training rectified flow models with oracle/conditional target on a 100-class ImageNet~\cite{deng2009imagenet} subset. Oracle supervision offers a slight advantage in early training (due to more accurate guidance on near-prior timesteps), while the later training dynamics largely overlap (as the oracle target collapses to the conditional target on most timesteps).}
\vspace{-4pt}
\label{fig:supp_oracle_training}
\end{figure}

\newpage

\subsection{Stage-aware Training Timestep Sampling}
\label{subsec:nonuniform_training}

\begin{table}[h!]
\centering
\scriptsize
\begin{tabular}{cccc}
\toprule
\textbf{Fraction of Stage 1 Training Steps} & 5\% & 10\% (uniform) & 20\% \\
\midrule
ImageNet gFID@5K $\downarrow$ & \textbf{23.69} & 24.28 & 25.34 \\
\bottomrule
\end{tabular}
\vspace{-4pt}
\caption{gFID convergence with stage-aware timestep sampling. We set the stage shift at $t=0.1$ and replace the uniform sampler with a piecewise-uniform distribution that controls the fraction of training steps allocated to each stage. Under uniform sampling, 10\% of timesteps fall in Stage~1 by default. Allocating more training steps to Stage~2 accelerates convergence, consistent with the observation that the benefits of
additional capacity and training compute are primarily reflected in the refinement stage (Sec.~\ref{subsec:learning_differs}). Evaluated on LightningDiT-B at 160K steps, no CFG.}
\vspace{-8pt}
\label{tab:supp_stage1_fraction}
\end{table}

\vspace{135pt}
\section{Findings on Flux.1[dev]}
\label{sec:flux_findings}

\subsection{Timestep-shifted Sampling on Flux.1[dev]}
\label{subsec:flux_inference}

\begin{figure}[h!]
\centering
\vspace{-12pt}
\includegraphics[width=0.98\linewidth]{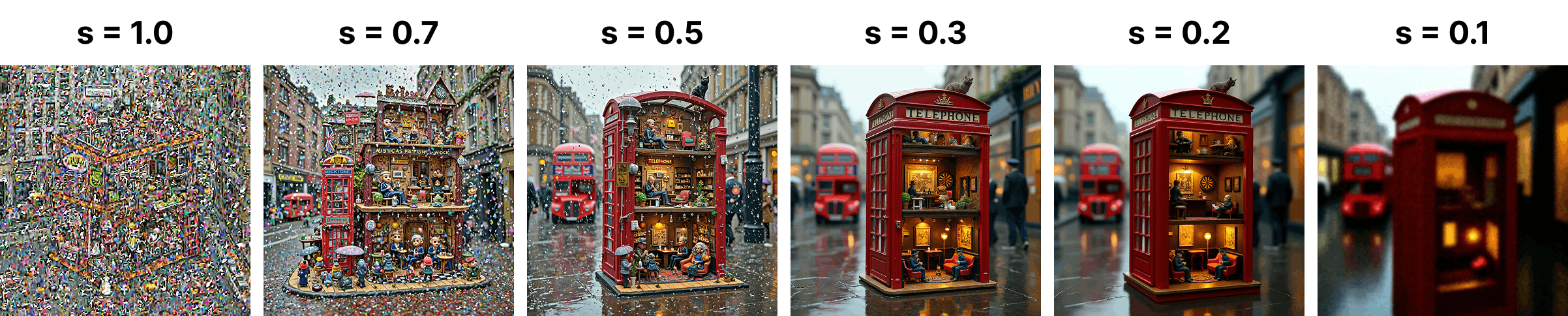}
\vspace{-4pt}
\caption{Flux.1[dev] generations under varying timestep shifts.}
\label{fig:supp_timestep_shift_flux}
\vspace{-8pt}
\end{figure}

Based on the oracle velocity, higher-resolution models like Flux/SD3 exhibit a faster concentration of the top-1 posterior, leading to a shorter navigation interval. Hence, these models become more sensitive to timestep allocation (during both training and inference) due to the condensed navigation stage. On a Flux.1[dev] model, \textit{a properly timestep-shifted sampling schedule (i.e., allocating slightly more navigation steps) improves both content fidelity and visual aesthetics} (Tab.~\ref{tab:supp_timestep_shift}), while an inappropriate shift can lead to corrupted content (Fig.~\ref{fig:supp_timestep_shift_flux}).

\begin{table}[h!]
\centering
\scriptsize
\begin{tabular}{lcccccc}
\toprule
\textbf{Timestep Shift $s$} & 1.0 & 0.7 & 0.5 & 0.3 & 0.2 & 0.1 \\
\midrule
CLIP-Score $\uparrow$ & 22.02 & 27.69 & 28.84 & \textbf{29.16} & 27.26 & 21.10 \\
LAION-Aesthetics $\uparrow$ & 4.96 & 6.37 & 6.90 & \textbf{7.22} & 7.11 &  4.39 \\
\bottomrule
\end{tabular}
\vspace{-4pt}
\caption{Timestep-shifted sampling in Flux.1[dev]: slightly more navigation steps (i.e., smaller $s$) improves image quality.}
\label{tab:supp_timestep_shift}
\end{table}

\onecolumn

\subsection{Analysis of Two-Stage Behaviors in Flux.1[dev]}
\label{subsec:flux_two_stage}

\begin{figure*}[h!]
\centering
\includegraphics[width=0.98\linewidth]{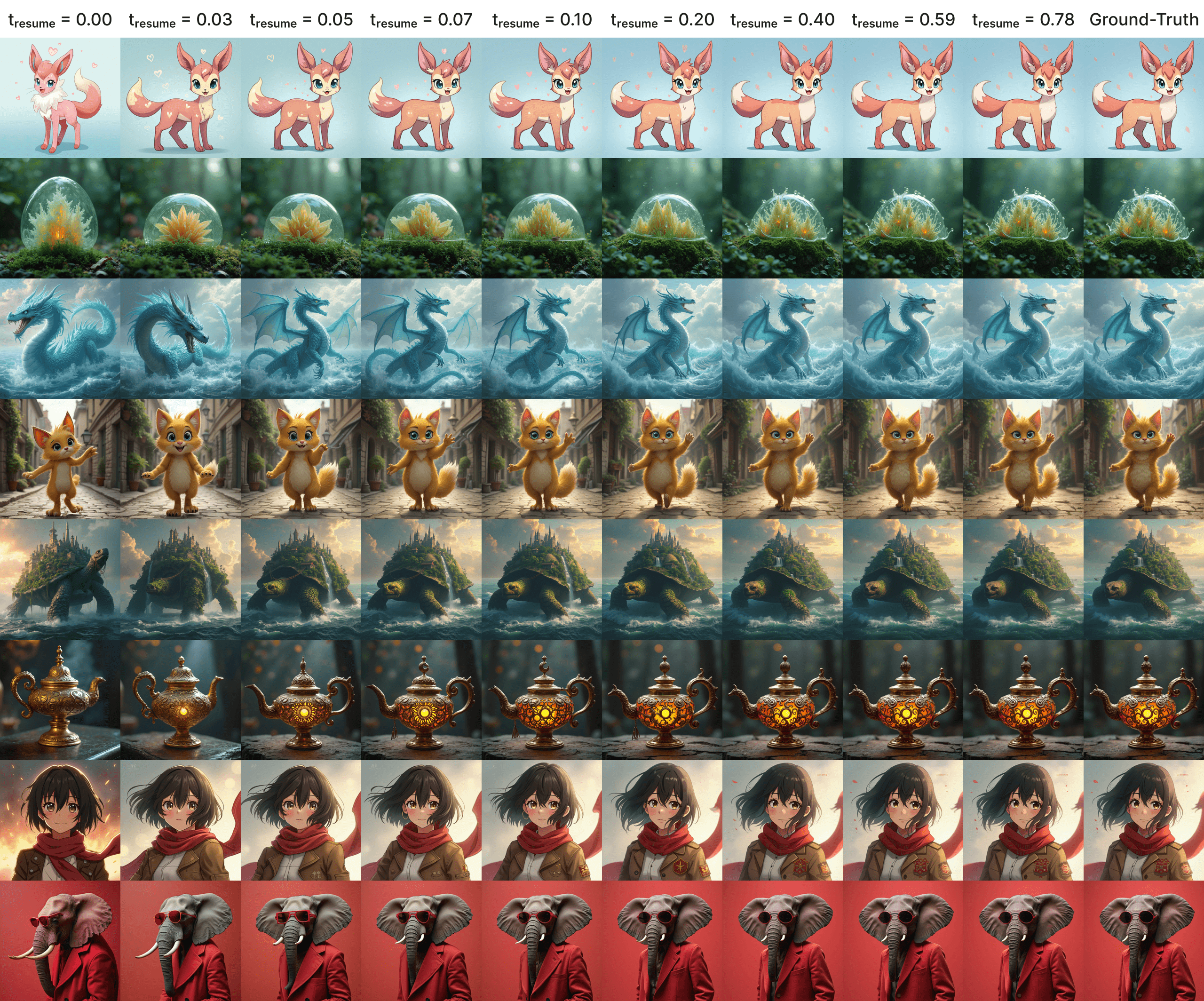}
\vspace{-4pt}
\caption{Qualitative illustration of two-stage behavior in \texttt{Flux.1[dev]}~\cite{flux2023}. Specifically, we first generate a reference latent $z_{\text{gt}}$ via text-to-image sampling. Then, we re-noise it by interpolating with Gaussian noise at a chosen $t_{\text{resume}}$ and resume sampling. Owing to Flux’s higher-dimensional latent space, the stage transition appears earlier than models trained on $256^2$ ImageNet data, and the model can reliably recover nearly identical images even after $\sim\!90\%$ noise corruption. We also note that Flux employs a non-uniform, resolution-aware timestep schedule and a different time convention; all $t_{\text{resume}}$ values shown here are converted to our convention for consistency.}
\vspace{-8pt}
\label{fig:supp_flux_resume}
\end{figure*}

\newpage
\section{Additional Qualitative Results}
\label{sec:additional_qual}

\begin{figure*}[h!]
\centering
\includegraphics[width=0.98\linewidth]{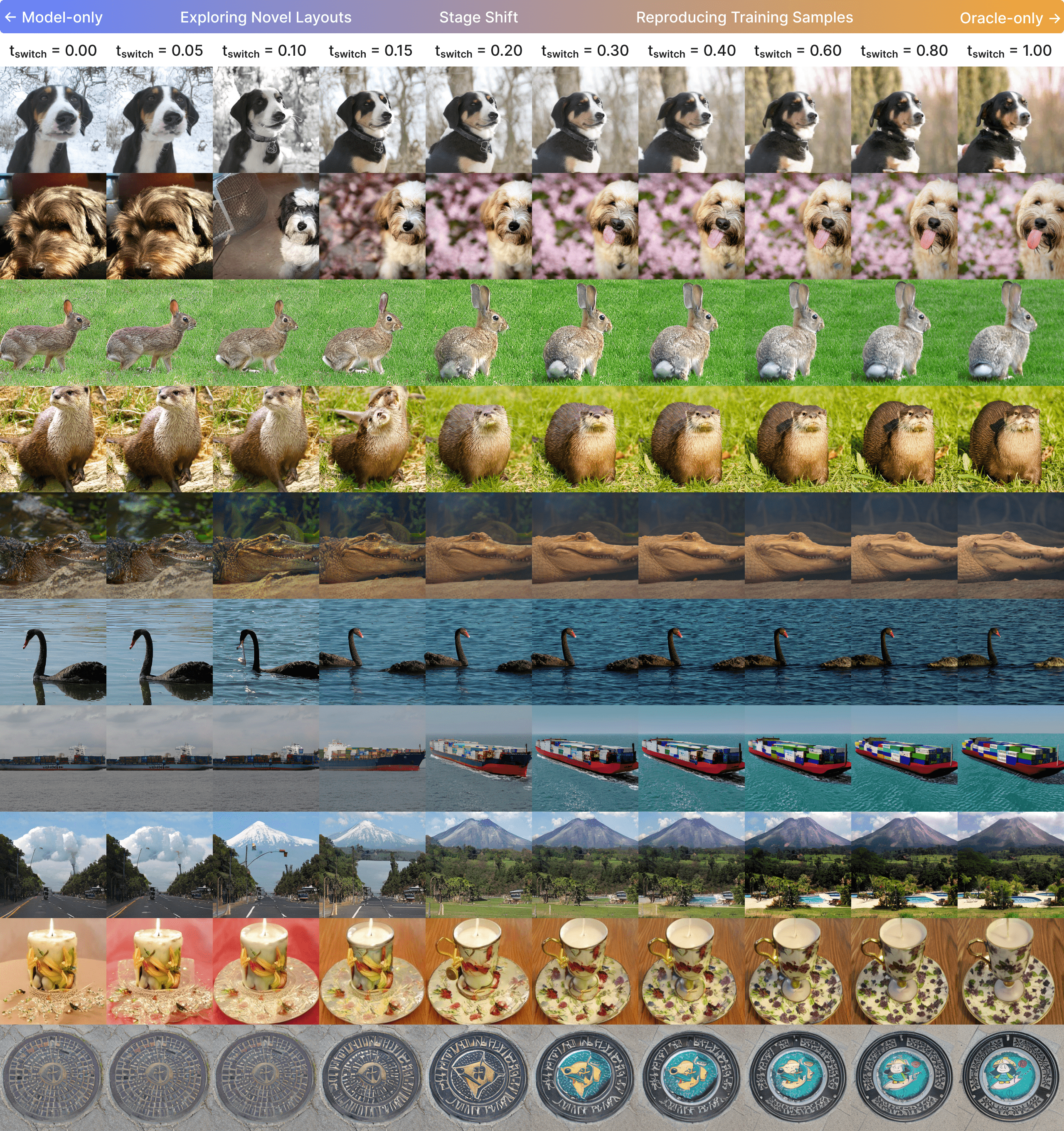}
\vspace{-4pt}
\caption{Mixed sampling results with switch point $t_{\text{switch}}$. Oracle $u_t^*$ is used before $t_{\text{switch}}$ and LightningDiT-XL/1~\cite{yao2025reconstruction} afterward. Overall, early switching yields diverse novel outputs (\textcolor{generalization}{generalization}), while late switching reproduces training samples (\textcolor{memorization}{memorization}). Despite minor variations across sampling trajectories, the empirical stage transition (i.e., reverting to training-like layouts) emerges around $t = 0.2$, slightly lagging behind the shift in the training target. Zoom in for the best view. Better view with color.}
\vspace{-8pt}
\label{fig:supp_mix_gen}
\end{figure*}

\begin{figure*}[h!]
\centering
\includegraphics[width=0.98\linewidth]{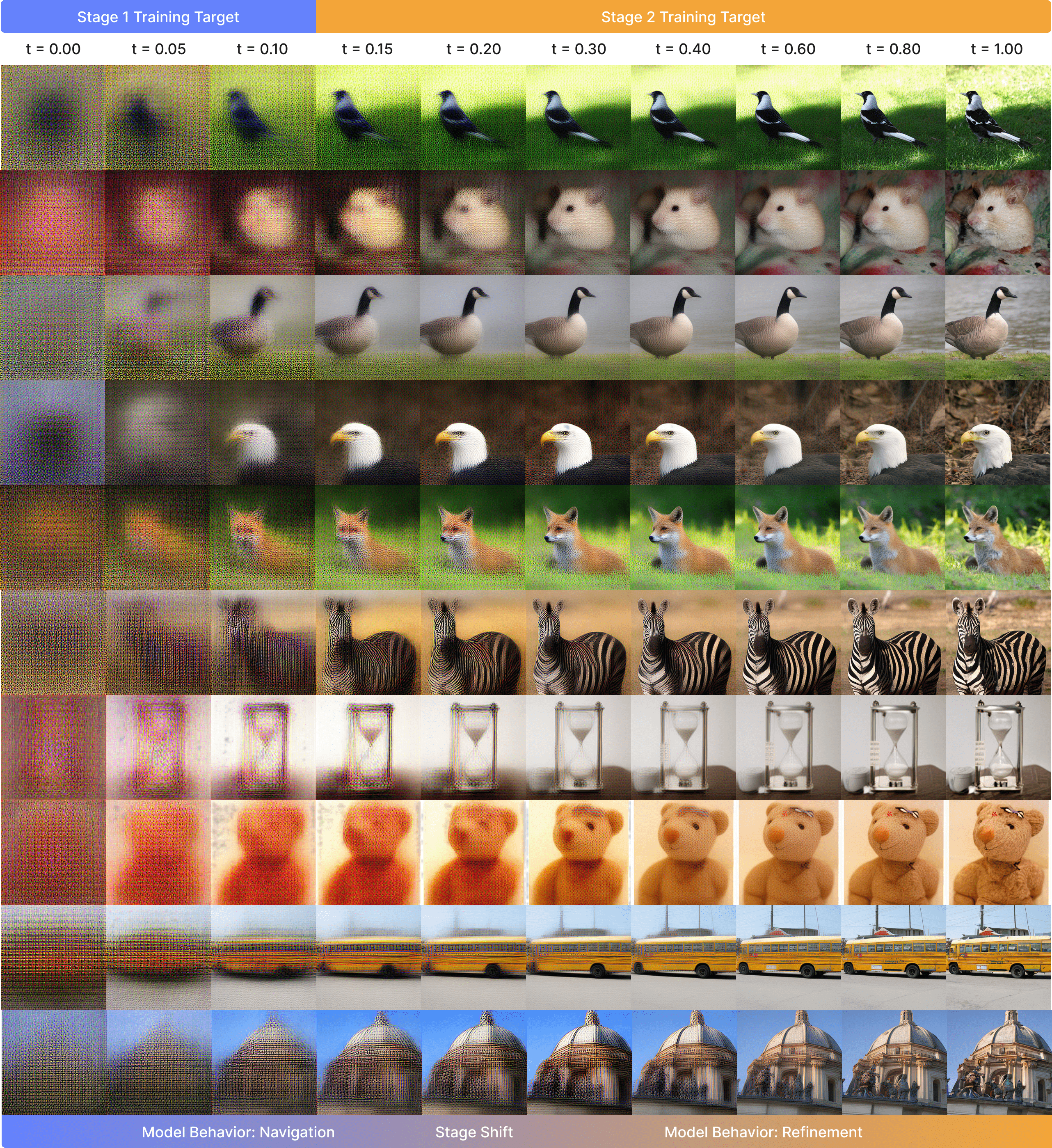}
\vspace{-4pt}
\caption{Intermediate predictions of a LightningDiT-XL/1~\cite{yao2025reconstruction} model across timesteps. Overall, early stages primarily \textcolor{generalization}{navigate} global layout via smoothed predictions, while later stages \textcolor{memorization}{refine} fine-grained details. Zoom in for the best view. Better view with color.}
\vspace{-8pt}
\label{fig:supp_interm_pred}
\end{figure*}